%% file: main.tex
\documentclass[11pt]{article}

\usepackage[margin=1in]{geometry}
\usepackage{xcolor}
\usepackage{authblk}

\usepackage[numbers,sort&compress]{natbib}

\input{math_commands.tex}

\usepackage{hyperref}
\usepackage{url}
\usepackage{booktabs} %
\usepackage{multirow} %
\usepackage{amsmath}
\usepackage{amssymb}
\usepackage{mathtools}
\usepackage{amsthm}
\usepackage{subcaption}
\usepackage{graphicx}
\usepackage{pifont} %
\usepackage{comment}
\usepackage[textwidth=5em,disable]{todonotes}

\hypersetup{
    colorlinks=true,
    linkcolor=black,
    filecolor=magenta,
    urlcolor=blue,
    citecolor=black
}

\theoremstyle{plain}
\newtheorem{theorem}{Theorem}[section]
\newtheorem{proposition}[theorem]{Proposition}

\theoremstyle{definition}
\newtheorem{definition}[theorem]{Definition}

\theoremstyle{remark}

\renewcommand{\L}{\ensuremath{\mathcal{L}}}
\renewcommand{\D}{\ensuremath{\mathcal{D}}}

\begin{document}

\title{Learning in the Recurrent State: Gradient Descent with Linear Recurrent Networks}

\author[1]{Yudou Tian}
\author[1]{Neeraj Mohan Sushma}
\author[2]{Harshvardhan Mestha}
\author[3]{Nicolo Colombo}
\author[1]{David Kappel}
\author[3]{Anand Subramoney\thanks{Corresponding author: \texttt{anand.subramoney@rhul.ac.uk}}}
\affil[1]{Center for Cognitive Interaction Technology CITEC, Universit\"at Bielefeld, Germany}
\affil[2]{Department of Electrical and Electronics Engineering, Birla Institute of Technology and Science Pilani, India}
\affil[3]{Department of Computer Science, Royal Holloway, University of London, United Kingdom}
\date{}

\maketitle

\begin{abstract}
In-context learning lets a sequence model adapt to a new task from examples in its input. 
A prominent line of work shows how self-attention can be constructed to implement gradient descent on a linear predictor fit to the in-context examples during the forward pass.
State-space models (SSMs) and other linear recurrent networks (LRNNs) model sequences at linear time cost, but it is unclear how their recurrent update could carry out the same in-context gradient descent.
We introduce Gradient-based Recurrent In-context Learner (GRIL), a diagonal LRNN that factorizes a supervised gradient step into a short-window cross-product write and a multiplicative readout of the next query.
For linear regression, this construction accumulates the context gradient in a matrix state and applies it in a single forward pass, with $\mathcal{O}(f^2)$ learned degrees of freedom.
The same design extends to multi-step updates and cross-entropy classification, with a limited MLP-based extension to non-linear regression.
We show empirically that trained GRILs recover the behavior and parameters analytically predicted by the construction on synthetic ICL tasks. 
Furthermore, the same architecture can be extended and trained on general-purpose benchmarks, including Long Range Arena, language modeling and associative recall.
Together, these results establish windowed cross-product self-attention as a concrete inductive bias that lets LRNNs learn in context through gradient-descent-like updates, while remaining trainable on general-purpose tasks. %
\end{abstract}

\section{Introduction}

Transformers~\citep{vaswani_attention_2017} are the default architecture for large-scale sequence modeling. In-context learning (ICL), which lets a sequence model adapt to a new task from examples in its prompt without changing its weights, has been proposed as one source of the capabilities they acquire at scale~\citep{wei_larger_2023, lu_are_2023}.
This ability is useful for few-shot deployment and provides a concrete setting for studying how a fixed network can implement learning during a forward pass.
But the quadratic sequence cost of transformers continues to motivate the search for recurrent alternatives~\citep{gu_efficiently_2021,gu_mamba_2023,orvieto2023resurrecting,peng_eagle_2024}.
Linear recurrent networks (LRNNs) are attractive because they support linear-time inference and parallel training through associative scan~\citep{smith_simplified_2022}.
Recent large-scale recurrent models are competitive in language modeling and can exhibit ICL~\citep{dao_transformers_2024,de_griffin_2024,grazzi_is_2024}.

A leading mechanistic account of transformer ICL is that self-attention can simulate gradient descent on a task-specific predictor %
~\citep{von_oswald_transformers_2023,akyurek_-context_2024,ahn_2023_transformersa,mahankali_one_2023} (but see~\citep{fu_2024_transformers}).
Linear self-attention can also be written as a recurrent update~\citep{katharopoulos_transformers_2020}, which motivates a design question: how should an LRNN be biased so that a fixed recurrent update can perform task adaptation inside its state during a forward pass?
We address this question as architecture design, rather than a post-hoc explanation of a mechanism already present in LRNNs.
For multivariate linear regression with input and target dimension $f$, the linear predictor fit to the in-context examples has $\mathcal{O}(f^2)$ degrees of freedom.

\begin{figure*}[!t]
    \centering
    \includegraphics[width=0.9\textwidth]{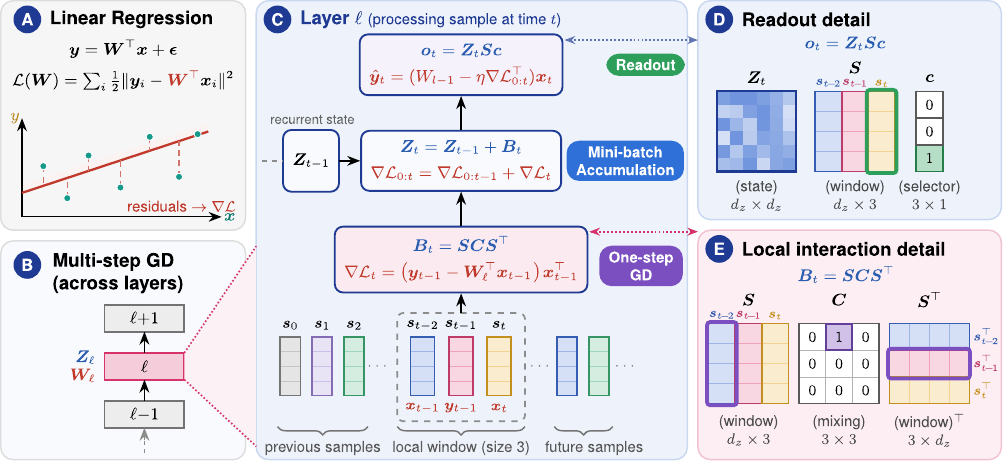}
    \small
    \caption{
    \textbf{GRIL architecture and its correspondence to gradient descent.}
    Blue denotes recurrent computations and red denotes their GD interpretation.
    To keep the figure simple we label the window tokens $\vs$ rather than separate query, key, and value projections; each $\vs$ corresponds to a query, key, or value depending on where it enters the windowed update and readout.
    (A)~The linear regression objective used to derive the GD construction.
    (B)~Stacking GRIL blocks across layers applies successive GD updates, with each layer corresponding to one update step (Appendix~\ref{appendix:multi}).
    (C)~Within each layer, a local window produces the recurrent update $\mB_t$, which is accumulated in the state $\mZ_t$ and used by the readout for the current query.
    (D)~The selector $\vc$ extracts the query token from the local window before multiplication by the accumulated state.
    (E)~The sparse mixing matrix $\mC$ selects the token pair used to form the outer-product update.
    }
    \label{fig:framework}
\end{figure*}

The generic gated-recurrent emulation of linear attention considered by~\citep{zucchet_2023_gated} can require $\mathcal{O}(f^4)$ parameters for the same setting.
This cost comes from how that construction parameterizes the update: the network holds the $f\times f$ predictor, and letting each element depend on all the others takes about $f^2\times f^2=f^4$ weights.
GRIL addresses this parameterization problem with a specific factorization, shown in Figure~\ref{fig:framework}: a length-three window exposes $(\vx_t, y_t, \vx_{t+1})$, the cross-product write accumulates $y_t\vx_t^T$ in the recurrent state, and the multiplicative readout applies this statistic to the next query.
For a zero-initialized linear predictor, one block therefore implements one mini-batch gradient-descent step using $\mathcal{O}(f^2)$ learned degrees of freedom; stacking blocks applies further steps to the predictor produced by the preceding layer.
GRIL is thus an explicit instance of mesa-optimization, in which the forward pass of a fixed network carries out an inner learning algorithm on the in-context data, built in by construction rather than emerging implicitly from training~\citep{von_oswald_uncovering_2023}.
Because each layer applies the same update, the multilayer construction can also be realized by weight tying and looping a single GRIL block.
We focus on a construction that performs mini-batch gradient descent rather than online, per-sample gradient descent.
An online update edits the predictor after every demonstration, which makes the result depend on the order in which demonstrations arrive, at odds with the exchangeability of in-context examples.
It also ties each update to a single-example gradient which is a high-variance estimate of the batch gradient~\citep{bottou_2018_optimization, goodfellow_2016_deep}.
Our experiments test whether training recovers this specified write-read factorization and whether the same block remains useful beyond the structured regression setting.

In summary, we:
\begin{itemize}
    \item formalize a diagonal LRNN block whose cross-offset write and next-query readout implement one mini-batch gradient step on the in-context least-squares objective;
    \item extend the construction to multi-step updates and cross-entropy classification, and study its limited MLP-based non-linear regression extension;
    \item give the construction with $\mathcal{O}(f^2)$ learned degrees of freedom for $f \times f$ regression, versus the $\mathcal{O}(f^4)$ parameterization of the generic gated-recurrent emulation of linear attention considered by~\citep{zucchet_2023_gated};
    \item show that trained GRILs recover the predicted gradient-descent dynamics and parameters on synthetic regression and classification ICL tasks;
    \item and evaluate the same architectural bias on Long Range Arena (LRA), language modeling and long-context associative recall (AR) benchmarks, showing that it remains trainable in standard sequence-model pipelines. %
\end{itemize}

\section{Related Work}
\paragraph{ICL as implicit optimization.}
A central line of work studies whether in-context learning implements a recognizable learning algorithm inside the forward pass.
Transformers can learn in context by emulating gradient descent on linear prediction tasks~\citep{von_oswald_transformers_2023}, and one layer of linear self-attention can implement an optimal one-step gradient update in a related setting~\citep{mahankali_one_2023}.
Other analyses identify algorithmic structure in linear-model ICL~\citep{akyurek_what_2022}, preconditioned gradient descent in transformers~\citep{ahn_2023_transformersa}, second-order convergence for in-context linear regression~\citep{fu_2024_transformers}, training dynamics for linear-attention ICL~\citep{zhang_training_2025}, and mesa-optimization algorithms inside trained transformers~\citep{von_oswald_uncovering_2023}. 
The broader idea of adaptation without weight updates predates modern ICL, appearing in fixed-weight learning rules~\citep{cotter_1990_fixedweight}, meta-learned recurrent systems~\citep{hochreiter_2001_learning,andrychowicz_learning_2016,wang_2016_learning}, and fast adaptation through neural dynamics alone~\citep{subramoney_2024_fast}.

The observation that a single layer can realize one gradient update raises the question of how multiple optimization steps can be implemented within a sequence model.
Looped Transformers address this through repeated shared-parameter updates that implement multi-step gradient descent in linear ICL settings~\citep{gatmiry_looped_2024,chen_bypassing_2024}, while MesaNet derives its computation from an in-context regression objective that is solved to local optimality at each timestep~\citep{von_oswald_mesanet_2026}.
Our focus is an explicit finite LRNN construction that realizes one mini-batch gradient-descent step over the available context while preserving the $\mathcal{O}(f^{2})$ task structure.
Appendix~\ref{appendix:looping} relates our construction to looped architectures.

\paragraph{Recurrent sequence models and learning-rule updates.}
Modern SSMs and LRNNs replace quadratic attention with recurrent state updates.
S4 uses structured state spaces for long sequences~\citep{gu_efficiently_2021}, and S5 simplifies state-space layers~\citep{smith_simplified_2022}; 
Mamba uses a diagonal recurrent matrix and makes the state update input-selective~\citep{gu_mamba_2023} and has also been shown to acquire in-context linear regression when trained directly on synthetic regression tasks~\citep{park2024mambalearnlearncomparative}. 
Eagle/Finch introduces data-dependent linear recurrences with diagonal, matrix-valued per-head states~\citep{peng_eagle_2024}.
Linear attention provides another recurrent formulation, where attention can be expressed through recurrent matrix-state updates~\citep{katharopoulos_transformers_2020}, while gated recurrent networks can emulate attention indirectly~\citep{zucchet_2023_gated}.
Diagonal, matrix-valued states naturally emerge in our GRIL construction, as do input-dependent, gated updates.
GRIL reads its query, key, and value projections from a short local window, relating it to the short convolutions used in these models; Appendix~\ref{appendix:shortconv} develops this relation and identifies the construction as a one-hot special case.

Fast weights memories provide a related view in which an internal weight state is modified during sequence processing~\citep{schmidhuber_learning_1992,ba_using_2016}.
\citet{schlag_linear_2021} makes this connection explicit for linear attention, interpreting its recurrent state as a fast-weight memory updated by additive key--value outer products. 
This fast-weight view extends naturally to recurrent architectures that replace additive writes with explicit learning-rule updates.
\citet{wang_testtime_2025} formalize this perspective, casting linear attention, delta-rule models, and test-time learners as a single family that solves an associative-memory regression problem at test time; GRIL sits in this family, with its distinguishing choice being a mini-batch least-squares update accumulated over the context rather than a per-token online write.
A parallel line of test-time-memorization models instantiates this view at scale: Titans learns a deep neural memory whose chunked update is a mini-batch gradient step with momentum and weight decay~\citep{behrouz_titans_2024}, Atlas optimizes the memory over the whole context rather than one token at a time~\citep{behrouz_atlas_2025}, and Miras organizes the choices of memory objective and retention into a common design space~\citep{behrouz_miras_2025}.
These models learn a self-supervised associative memory at scale, while GRIL gives an explicit construction for a supervised in-context regression objective with $\mathcal{O}(f^2)$ parameters and a verified correspondence to gradient descent.
Test-Time Training makes this view explicit by treating the recurrent state as a self-supervised learner updated by gradient descent during the forward pass~\citep{sun_learning_2024}.
Its update is still per-token and online, driven by a self-supervised reconstruction objective on the current token, so it shares the online-versus-mini-batch contrast we draw against the delta-rule family below.

DeltaNet applies delta-rule corrections to its recurrent state~\citep{yang_parallelizing_2024}, and Longhorn derives its state-space transition from an implicit online-regression objective~\citep{liu_longhorn_2024}.
Gated DeltaNet extends the delta-rule family with gated input processing that introduces cross-product interactions before the recurrent write~\citep{yang_gated_2025}, while later variants add separate erase and write gates~\citep{hatamizadeh_gdn2_2026} and curvature-aware preconditioning~\citep{tumma_preconditioned_2026}.
These models share GRIL's matrix-valued state and outer-product-like writes, but compute each update from the current token: DeltaNet applies a per-token delta-rule correction, and Longhorn derives its per-token update from an implicit online-regression objective on the current key--value association.
These per-token updates modify the fast-weight representation online, whereas GRIL accumulates supervised gradient statistics over the context seen so far and defers a single mini-batch update to the readout on the next query.
Appendix~\ref{appendix:longhorn-perspective} develops this contrast formally against Longhorn's online-regression update (Appendix~\ref{appendix:longhorn-view}) and reports matched online versus mini-batch GD and DeltaNet/Gated DeltaNet comparisons.

\section{Preliminaries}\label{sec:background}

A linear recurrent network processes a sequence of tokens $\vs_1,\ldots,\vs_T$, collected as the rows of $\mS \in \R^{T \times f}$, where $T$ is the context length and $f$ is the token feature dimension, by updating a recurrent state $\mZ_t \in \R^{d \times m}$ and emitting an output $\vo_t$. Here $d$ and $m$ are the value and key/state dimensions used by the recurrent representation. We write this generic update and readouts as
\begin{align}
    \mZ_t &= \mA_t(\mS') * \mZ_{t-1} + \mB_t(\mS')\,, \label{eq:ssm} \\
    \vo_t &=  \mZ_t\, \mU_t(\mS')\,, \qquad \vo'_t = \vo_t * \sigma (\mU_1 \vo_t )\,. \label{eq:ssmout}
\end{align}
Here $\mS' \subset \mS$ is a length-$\tau$ subsequence (usually one token), $\mA_t,\mB_t: \R^{\tau \times f} \rightarrow \R^{d \times m}$ are the state-transition and input maps, $\mU_t$ is a compatible readout map, $\mU_1$ is an optional output-gating map, $\sigma$ is applied elementwise, and $*$ denotes the relevant elementwise or matrix product.

\begin{table}[tbp]
\centering
\small
\caption{Matrix-state recurrent updates in related models.
For compactness, $\vk_t,\vv_t,\vq_t$ denote key, value, and query projections; $\alpha_t$, $\beta_t$, and $\Delta_t$ denote the state-decay, delta-rule, and Longhorn update coefficients, respectively. 
For GRIL, $C_{ij}$ denotes the window-mixing coefficient (see Eq.~\ref{eq:window-cross-product}).
}
\label{tab:recurrent-comparison}
\setlength{\tabcolsep}{3pt}
\begin{tabular}{lll}
\toprule
Model & State update & Readout \\
\midrule
Linear attention
& $\mZ_{t-1}+\vv_t\vk_t^T$
& $\mZ_t\vq_t$ \\

(recurrent form)
& 
&  \\

DeltaNet
& $\mZ_{t-1}+\beta_t(\vv_t-\mZ_{t-1}\vk_t)\vk_t^T$
& $\mZ_t\vq_t$ \\

Gated DeltaNet
& $\alpha_t\mZ_{t-1}
 +\beta_t(\vv_t-\alpha_t\mZ_{t-1}\vk_t)\vk_t^T$
& $\mZ_t\vq_t$ \\

Longhorn
& $\mZ_{t-1}
 +\Delta_t(\vx_t-\mZ_{t-1}\vk_t)\vk_t^T$
& $\mZ_t\vq_t$ \\

GRIL
& $\mZ_{t-1}+\sum_{i,j}C_{ij}\vv_i\vk_j^T$
& $\mZ_t\vq_t$ \\
\bottomrule
\end{tabular}
\end{table}

\section{An Inductive bias for ICL in LRNNs}
Figure~\ref{fig:framework} provides an overview of the mechanism studied in this section. 
We first consider a linear regression problem, where training examples are presented sequentially as context. 
In order to perform mini-batch gradient descent inside the recurrent dynamics, the network must both accumulate the gradient contributed by the given samples and use that to provide a prediction for the next query. 

\subsection{Warm-up: Scalar regression}\label{sec:1dregression}

We first provide a construction for the simple case of scalar linear regression.

For scalar regression, let $\D=\{(\vx_i,y_i)\}_{i=1}^N$ be a dataset with inputs $\vx_i\in\R^f$ and scalar targets $y_i\in\R$ generated by the task model $y=\vw^T\vx$,
where $\vw \in \R^{f}$ is the task parameter. The least-squares objective is
\begin{align}
    \L(\D; \vw) = \frac{1}{2N} \sum_{i=1}^N \left(\vw^T \vx_i - y_i \right)^2 \,. \label{eq:1dloss}
\end{align}
One step of mini-batch gradient descent with learning rate $\eta$ gives $\vw_1 = \vw_0 - \eta\,\nabla_{\vw}\L(\D_{1:t};\vw_0)$, where $\D_{1:t}$ denotes the first $t$ examples in $\D$ and
\begin{align*}
    \nabla_{\vw}\L(\D_{1:t};\vw_0)
    = \frac{1}{t}\sum_{i=1}^t
    \left(\vw_0^T \vx_i-y_i\right)\vx_i \,.
\end{align*}
Because this update subtracts the gradient, it is the negative gradient that the recurrent state accumulates,
\begin{equation}
    \vz_t = \mI \, \vz_{t-1} + (y_t - \vw_0^T \vx_t) \, \vx_t \,, \label{eq:1dlrnn}
\end{equation}
so that $\vz_t = -t\,\nabla_{\vw}\L(\D_{1:t};\vw_0)$ under the initialization $\vz_0=\vzero$, which corresponds to $\vw_0=\vzero$.
The readout applies this state to the next query and returns the prediction of the updated parameter,
\begin{align}
    \hat{y}_{t+1} = \vw_1^T \vx_{t+1} = \frac{\eta}{t} \vz_t^T \, \vx_{t+1} \,. \label{eq:predy}
\end{align}
Without a sliding window, $y_t$ and $\vx_{t+1}$ are split across different time steps, which prevents the gradient update in Eq.~\ref{eq:1dlrnn} and the prediction readout in Eq.~\ref{eq:predy} from being completed within a single recurrent cycle.
The scalar warm-up avoids this by packaging the three factors into one constructed token, as its experiments do (Appendix~\ref{appendix:expt-1d-regression}), and the sliding window of the next section removes that packaging by reading the factors off the interleaved sequence.
Appendix~\ref{appendix:1dlinreg} gives the full algebraic derivation.

\subsection{One-step GD for multivariate linear regression}
\label{sec:ndregression}

We now generalize the construction to vector-valued linear regression with predictor $\hat{\vy}=\mW^T\vx$.
Without loss of generality we take $\vx, \vy \in \R^f$, since unequal dimensions can be matched by appropriate embeddings\footnote{For example, dimensions $k$ and $l$ can be embedded into dimension $k+l=f$ by concatenating zero vectors, or both can be mapped linearly into dimension $f$.}.
The system is then $f$ coupled scalar regressions, one per coordinate of $\vy$.
\begin{proposition}
    Given a diagonal linear recurrent layer with a sliding window of size 3 and stride 2, and tokens $\vs_{2j-1} = \vx_j$ and $\vs_{2j} = \vy_j$ for $j = 1, \ldots, N$ drawn from a linear model, one can construct recurrent matrix $\mA$, input $\mB$, and output matrix $\mU$ such that each recurrent step produces $\hat{\vy}_{j+1} = \mW_1^T \vx_{j+1}$, where $\mW_1 = \mW_0 - \eta \nabla_\mW \L(\D_{1:j}; \mW_0)$ is one step of gradient descent from $\mW_0 = \vzero$.
    The test input $\vx_{N+1}$ is contained in token $\vs_{2N+1}$, and produces the test prediction $\hat{\vy}_{N+1}$.
\end{proposition}

Following Eq.~\ref{eq:1dlrnn}, we write the LRNN update and its output as
\begin{align}
    \mZ_t &= \mZ_{t-1} + \vy_t \, \vx_t^T \,, \label{eq:ndlrnn} \\
    \vo_t &= \beta \mZ_t \, \vx_{t+1} \,, \label{eq:ndlrnn-output}
\end{align}
where $\mZ_t$ corresponds to the accumulated update for a linear predictor $\mW \in \R^{f \times f}$ fit to the in-context examples, and we assume, for simplicity, that the initial predictor is $\mW_0=\vzero$\footnote{The more general case is discussed in Appendix~\ref{appendix:multi}.}, $\beta$ is the scalar readout coefficient; for exact mini-batch GD over the first $t$ examples, $\beta=\eta/t$.
In the synthetic ICL tasks the context length is fixed per task, so $\beta=\eta/N$ is a constant and the $1/t$ form is a property of the exact construction rather than a runtime rescaling the trained model must implement.
In the general-purpose %
models, where the effective context length varies, the pre-normalization layers absorb the growth of $\mZ_t$ with $t$, so a learned static $\beta$ suffices.

To put this update in the form of Eq.~\ref{eq:ssm}, we let the input sequence contain the training dataset of the linear regression task: $\vs_1, \vs_2, \ldots$, where $\vs_{2j-1} = \vx_j$ and $\vs_{2j} = \vy_j$, so that $\vs_{2j+1} = \vx_{j+1}$.
At each step, Eq.~\ref{eq:ndlrnn} and Eq.~\ref{eq:ndlrnn-output} require $\vx_t$, $\vy_t$, and $\vx_\tpone$.
We therefore use a length-three sliding window with stride two.
Let $\mV_t,\mK_t,\mQ_t\in\R^{f\times 3}$ be the value, key, and query embeddings of this window, generally three distinct learned projections of the window tokens as in attention. Each shares the same form: writing $\mE_t$ for any of $\mV_t,\mK_t,\mQ_t$, it stacks the corresponding projections of the three window tokens as columns,
\begin{align}
    \mE_t &= \begin{bmatrix}
        \vdots & \vdots & \vdots \\
        \vs_{2t-1} & \vs_{2t} & \vs_{2t+1} \\
        \vdots & \vdots & \vdots
    \end{bmatrix}\,, \qquad \mE_t \in \{\mV_t,\mK_t,\mQ_t\}\,, \label{eq:context}
\end{align}
where each column $\vs$ is the value, key, or query projection of the corresponding window token, according to which embedding $\mE_t$ denotes, so the three embeddings share this structure but take different values. In the construction all three projections are the identity, so $\mV_t = \mK_t = \mQ_t$ and the columns are the raw tokens $(\vs_{2t-1},\vs_{2t},\vs_{2t+1}) = (\vx_t,\vy_t,\vx_\tpone)$.

The corresponding windowed cross-product is
\begin{align}
    \mV_t \mC \mK_t^T = \sum_i \sum_j C_{ij}\, \vv_i \vk_j^T\,. \label{eq:window-cross-product}
\end{align}
Here $\mC \in \R^{w \times w}$ is the window-mixing matrix, and $\vv_i=[\mV_t]_{:,i}$ and $\vk_j=[\mK_t]_{:,j}$ are the columns of the value and key embeddings. This resembles the single-token outer product of linear self-attention (Table~\ref{tab:recurrent-comparison}), except that it forms a weighted sum of outer products over the entire window. Appendix~\ref{appendix:shortconv} gives an alternative formulation of this update based on convolutions that is similar to that used in existing SSMs.
If $\mC = \left(\begin{smallmatrix}0 \\ 1 \\ 0\end{smallmatrix}\right) \left(\begin{smallmatrix}1 & 0 & 0\end{smallmatrix}\right) = \left[\begin{smallmatrix}0 & 0 & 0 \\ 1 & 0 & 0 \\ 0 & 0 & 0\end{smallmatrix}\right]$ is chosen, substituting Eq.~\ref{eq:window-cross-product} as the input map in the generic update Eq.~\ref{eq:ssm} yields the update term $\vy_t \vx_t^T$ of Eq.~\ref{eq:ndlrnn}.

The LRNN can therefore be written in the generic form of Eq.~\ref{eq:ssm} by taking $\mA=\mI$ and Eq.~\ref{eq:window-cross-product} as the input map.

For the readout, if $\vc = \left(\begin{smallmatrix} 0 \\ 0 \\ 1 \end{smallmatrix}\right)$ is chosen to select the next query, the query vector $\vq_t := \mQ_t\vc$ equals $\vx_\tpone$, and the readout $\mZ_t\vq_t$ produces $\hvy_\tpone$ as in Eq.~\ref{eq:ndlrnn-output}.
This output matches the first readout in Eq.~\ref{eq:ssmout} with $\mU=\beta\mQ_t \vc$. %
Appendix~\ref{appendix:ndlinreg} gives the full derivation.

\begin{definition}[GRIL block]
A GRIL block is an LRNN whose recurrent update receives value, key, and query embeddings $\mV_t,\mK_t,\mQ_t\in\R^{f\times w}$ of a local window, each stacking the corresponding token-wise projections over the window as columns, in the same form as Eq.~\ref{eq:context}.
The block forms
\[
    \mZ_t = \mA \odot \mZ_{t-1} + \mV_t \mC \mK_t^T, \qquad
    \vo_t = \mZ_t \vq_t .
\]
Here $\mA$ is an element-wise recurrent gate (applied via the Hadamard product $\odot$) with the same shape as $\mZ_t$, $\mC\in\R^{w\times w}$ contains the cross-product weights, and $\vc\in\R^w$. %
When training models with this architecture, the learnable objects are the value, key, and query embeddings, $\mC$, $\vc$, and the diagonal recurrent/readout scalars; the constructive proofs fix these objects to the values specified above. 
k\end{definition}
The construction lets the LRNN perform gradient descent on an arbitrary-dimensional linear prediction task.
At constant window size it uses a matrix recurrent state and diagonal gates with $\mathcal{O}(f^2)$ learned degrees of freedom. 
On ICL tasks we train these parameters from a random initialization under mean-squared error.

The recurrent state is matrix-valued, as in Eq.~\ref{eq:ssm} and in many recent LRNN and SSM variants~\citep{dao_transformers_2024}; the distinction is that the recurrent input sums outer products over a local window of size $w$.
Setting $w=1$ reduces Eq.~\ref{eq:window-cross-product} to the single-token outer product $\vv(\vs_t)\vk(\vs_t)^T$ with readout $\mZ_t\vq(\vs_t)$, recovering the recurrent form of linear self-attention (Table~\ref{tab:recurrent-comparison}; Appendix~\ref{appendix:lsa}). The full GRIL construction uses $w=3$, exposing $(\vx_t,\vy_t)$ for gradient accumulation and the next query $\vx_{t+1}$ for applying the resulting update at readout.

Because the state first accumulates the gradient contributions from the in-context examples and the readout then applies their sum to the query, the construction performs a mini-batch gradient step.
Accumulating first keeps the state as an exact summary of the full-context gradient, so a readout reproduces the update a batch learner would compute.
Appendix~\ref{appendix:online-vs-minibatch} compares the two update rules directly under a matched number of updates.

\subsection{Beyond one-step linear regression}
\label{sec:nonlinear}

\paragraph{Multi-step gradient descent:}
The construction presented above extends to multi-step gradient descent by stacking GRIL layers, with each layer applying one gradient step to the predictor produced by the preceding layer (Appendix~\ref{appendix:multi}).
For the first $t$ in-context examples, define
$\mZ_t=\sum_{i\leq t}\vy_i\vx_i^T$ and
$\tilde{\mZ}_t=\sum_{i\leq t}\vx_i\vx_i^T$.
Each layer uses two cross-product heads, one accumulating $\mZ_t$ and the other $\tilde{\mZ}_t$, extending the single-state one-step block with a second window-mixing matrix $\tilde{\mC}$.
Starting from $\mW_0=\vzero$, layer $l$ implements
\begin{align*}
    \mW_l = \mW_{l-1} - \frac{\eta}{t}
    \left( \tilde{\mZ}_t\mW_{l-1} - \mZ_t^T \right)
\end{align*}
which is the gradient-descent update evaluated at the predictor $\mW_{l-1}$ produced by the preceding layer.
The recurrent statistics $\mZ_t$ and $\tilde{\mZ}_t$ depend only on the in-context examples and are therefore unchanged across layers, while the task-specific predictor $\mW_{l-1}$ is what changes from one layer to the next.
This predictor is not transmitted between layers as an explicit matrix; it enters only multiplicatively at the readout, where the two data-dependent recurrent states are combined to form the output.

Since every layer applies the same update to the same in-context data, the multilayer construction can be written as a looped model by tying the GRIL parameters and repeatedly applying a single block.
At each iteration, the original context is supplied again, while only the current task-specific predictor is propagated through the loop.
Thus, $L$ iterations of the tied block implement $L$ successive gradient steps.

\paragraph{Non-linear regression and classification:}
A limited extension to non-linear prediction is possible by adding MLP layers to GRIL.
The recurrent layer still accumulates the linear sufficient statistics, while the MLP maps them to quantities corresponding approximately to the gradient of a non-linear predictor. The mapping is only approximate because the exact non-linear gradient applies the non-linearity inside the per-example sum, so it depends on sufficient statistics (such as $\sum_i \vx_i$) that the linear state does not retain. Appendix~\ref{appendix:nonlinear} gives the non-linear construction.
Cross-entropy gradients retain the same outer-product structure, giving an exact GRIL construction when the prediction error is available in the local context (Appendix~\ref{appendix:celoss}). Our classification experiments instead use the non-linear construction above, with sigmoid or softmax applied at the output, and provide an approximate functional test of this extension (Appendix~\ref{appendix:clsexp}).

\paragraph{Regularization terms in the loss:}
Because the recurrent layers only accumulate gradients, gradient accumulation can be separated from the step that applies the update.
This gives a practical advantage: any input-independent regularization term, such as an L2 penalty, can be added without changing the recurrence.
See Appendix~\ref{appendix:regularisation} for details.

\section{Experiments}
\label{sec:exp}

We evaluate whether trained GRILs recover the dynamics predicted by the construction and whether the same inductive bias remains useful. Every GRIL reported below is trained from a random initialization.
The synthetic regression, ablation and classification experiments use randomly generated in-context datasets.
LRA, WikiText-103 and AR then run the same block in a standard sequence-model architecture.

\subsection{Synthetic regression}

\begin{figure*}[htbp]
    \centering
    \includegraphics[width=0.80\textwidth]{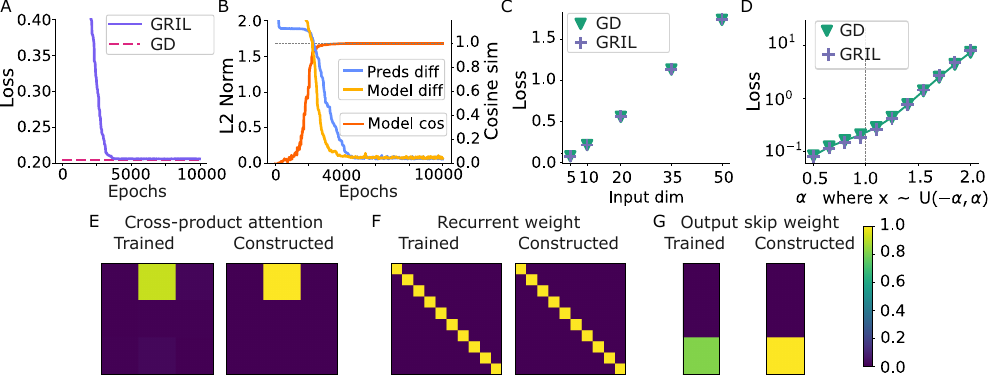}
    \small
    \caption{\textbf{A trained single-layer GRIL recovers both the predictions and the parameter structure of the one-step GD solution on multivariate regression.}
    GD denotes the exact one-step gradient-descent update for the linear regression objective.
    \textbf{A}: training loss.
    \textbf{B}: cosine similarity and $L_2$ distance between GD and GRIL, for parameters and for predictions.
    \textbf{C}: loss across input dimensions $f$.
    \textbf{D}: loss on inputs sampled at different scales from the training distribution.
    \textbf{E--G}: trained versus constructed input-gating weights, recurrent parameters, and skip-connection weights.
    All losses and similarities are means over 5 seeds.
    }
    \label{fig:lossnmatrices}
\end{figure*}

The clearest test of the theory is the multivariate regression setting from Section~\ref{sec:ndregression}, where the sequence is presented as $\vx_1,\vy_1,\vx_2,\vy_2,\ldots$ and each recurrent step has access to the local window $(\vx_t,\vy_t,\vx_{t+1})$.
The final query token contains the test input only, and its target is held out for the loss.
Appendix~\ref{appendix:expt-nd-regression} gives the corresponding data-generation details.
We compare a trained GRIL against the explicit gradient-descent construction using prediction error, sensitivity alignment, robustness to distribution shift, and direct parameter comparison.

Figure~\ref{fig:lossnmatrices} shows that the trained model closely matches the one-step gradient-descent construction across output-space metrics and that the learned recurrent, input, and skip parameters align with the theoretical construction beyond merely reproducing the right predictions.
This is the main empirical support that the cross-product update identifies a solution naturally found by optimization.

\begin{figure*}[htbp]
    \centering
    \includegraphics[width=0.80\textwidth]{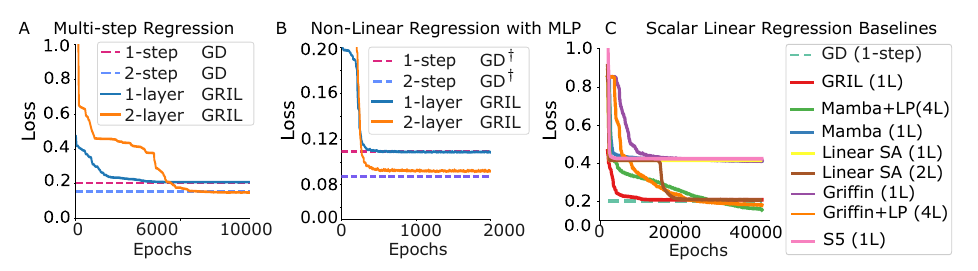}
    \small
    \caption{\textbf{Beyond one-step regression.}
    The same local cross-product mechanism supports multi-step GD, scalar regression, and a limited MLP-based non-linear extension.
    \textbf{A}: multi-step regression.
    \textbf{B}: 
    GD$^\dagger$ is the non-linear extension of the GD construction (Appendix~\ref{appendix:nonlinear}).
    \textbf{C}: 
    LP denotes a linear projection; Linear SA is linear self-attention with one or two layers.
    }
    \label{fig:result-regression-multi-layer}
\end{figure*}

Figure~\ref{fig:result-regression-multi-layer} shows that stacking GRIL layers reproduces multi-step GD (A), an MLP gives a limited non-linear extension (B), and one-layer GRIL outperforms one-layer generic SSM baselines on scalar regression (C); the two-layer transformer is the closest non-recurrent baseline.
In B, GD$^\dagger$ fixes the recurrent GRIL parameters and trains only the auxiliary MLP, testing whether it can transform the GD-like recurrent statistic (Appendix~\ref{appendix:expt-nonlinear}).
Two-layer parameter visualizations are in Figure~\ref{appendix:fig:2layer-weights}, the multi-step protocol is extended to three and four layers in Appendix~\ref{sec:multi-layer-depth}, and the multivariate comparison and recurrent-baseline configurations are in Figure~\ref{appendix:fig:nd-comparison} and Table~\ref{appendix:model_comparison}; the scalar setup is in Appendix~\ref{appendix:expt-1d-regression}.

\subsection{Ablation studies}
We ablate the GRIL components, with full scalar and multivariate tables in Tables~\ref{appendix:ablation-1d} and~\ref{appendix:ablation_study}.
Removing either the windowed write path or the multiplicative readout approximately doubles the loss, while the full GRIL matches the explicit GD construction.
We further compare against a one-step Newton--Raphson update (Table~\ref{appendix:second-order-methods}) and matched DeltaNet and Gated DeltaNet baselines (Appendix~\ref{sec:deltanet-baselines}); these reach a lower loss than one-step GD, but the task tests \emph{which} one-step algorithm a model implements, so any predictor that beats GD implements something other than one-step GD.
GRIL is the only model whose trained loss and parameters coincide with the analytic one-step GD solution, supporting the mechanistic reading that it recovers the first-order construction and weakening a generic stronger-optimizer explanation.

\subsection{Classification extension}

We test GRIL on the classification setup as well.
Figure~\ref{fig:classification-main} shows a representative result on a 3-class softmax task with sequence length 10.
Empirically, GRIL tracks GD$^\dagger$ on binary and multiclass synthetic tasks across sequence lengths and data scales, matching loss and directional sensitivity while sometimes reaching parameterizations that differ more than their predictions do. 
The GD-like signatures therefore persist under the classification non-linearity, though only with a learned non-linear feature map, so that the recurrent state accumulates the relevant classification update in feature space.
Appendix~\ref{appendix:classificationdata} describes how the classification labels are generated from the underlying linear task. 
Appendix~\ref{appendix:clsexp} gives the hyperparameter sweep (Table~\ref{tab:hyperparams}) and the remaining binary and length-100 results (Figures~\ref{fig:appendix:class-sigmoid-10}--\ref{fig:appendix:class-softmax-100}).

\subsection{General-purpose sequence tasks}
\label{sec:general-tasks}

\begin{table}[htbp]
\centering
\small
\caption{Long Range Arena test accuracy (\%). GRIL entries are means over three seeds. %
}
\label{tab:main-lra-results}
\setlength{\tabcolsep}{3pt}
\begin{tabular}{lccccc}
\toprule
\textbf{Model} & \textbf{ListOps} & \textbf{Text} & \textbf{Retrieval} & \textbf{Image} \\%
\midrule
S5 & \textbf{62.15} & 89.31 & \textbf{91.40} & 88.00 \\%
LRU & 60.20 & \textbf{89.40} & 89.90 & \textbf{89.00} \\%
Mamba & 38.0 & 83.0 & 72.1 & 69.8 \\%
GRIL & 61.58 & 86.64 & 87.68 & 71.98 \\%
\bottomrule
\end{tabular}
\end{table}

\begin{table}[htbp]
\centering
\small
\caption{WikiText-103 test perplexity under a matched trainer and training budget (GPT-2 tokenizer, non-overlapping $1024$-token windows; lower is better).
GRIL is matched in parameter count to the baselines Mamba-2~\citep{dao_transformers_2024}, TTT-Linear~\citep{sun_learning_2024}, and Gated DeltaNet-2~\citep{hatamizadeh_gdn2_2026}.}
\label{tab:main-wikitext-results}
\setlength{\tabcolsep}{3pt}
\begin{tabular}{lcc}
\toprule
\textbf{Model} & \textbf{Params} & \textbf{PPL} \\
\midrule
Gated DeltaNet-2 & 134.3M & \textbf{15.10} \\
Mamba-2 & 129.0M & 15.77 \\
TTT-Linear & 124.4M & 17.51 \\
GRIL & 127.8M & 16.77 \\
\bottomrule
\end{tabular}
\end{table}

\begin{table}[htbp]
\centering
\small
\caption{Test accuracy (\%) on long-context associative recall (sequence length $131$K, vocabulary $30$).
Hyena-S5, Mamba, and GRIL run under a matched outer LM architecture and training budget ($10$-seed means).
}
\label{tab:ar-131k-30}
\setlength{\tabcolsep}{3pt}
\begin{tabular}{lc}
\toprule
Model & Accuracy (\%) \\
\midrule
Transformer, FlashTransformer & \ding{55} \\
GSS, H3, AFT, RWKV            & $\le 2.3$ \\
\midrule
Hyena-Convolution~\citep{hyena} & $97.2$ \\
Hyena-S5                         & $97.2$ \\
Mamba                            & $99.5$ \\
\textbf{GRIL (ours)}             & $\mathbf{99.9}$ \\
\bottomrule
\end{tabular}
\end{table}

Next, we ask whether the GRIL block remains trainable and non-degenerate when the supervised $(\vx_t,\vy_t,\vx_{t+1})$ factors are no longer explicit in the data format, testing whether the gradient-descent bias transfers outside the structured synthetic tokenization.
We evaluate GRIL on LRA~\citep{Tay2021}, WikiText-103~\citep{merity2016pointer} and long-context AR~\citep{hyena}.
The write and read equations are unchanged from Section~\ref{sec:ndregression}, with the same windowed cross-product write and multiplicative readout (Figure~\ref{appendix:fig:framework} in Appendix~\ref{appendix:general-impl}).

On $131$K-token associative recall, GRIL achieves the highest accuracy among the models that solve the task: $99.9\%$ over 10 seeds, ahead of Mamba ($99.5\%$) and Hyena-S5 ($97.2\%$) (Table~\ref{tab:ar-131k-30}); per-seed results and baseline sources are in Appendix~\ref{sec:ar-associative-recall}.
On LRA, GRIL is within range of prior SSMs on ListOps, Text, and Retrieval and outperforms Mamba on all four reported tasks, while S5 and LRU remain stronger overall (Table~\ref{tab:main-lra-results}).
On WikiText-103, under the parameter-matched protocol GRIL outperforms TTT-Linear but trails Mamba-2 and Gated DeltaNet-2 (Table~\ref{tab:main-wikitext-results}).
Extended comparisons and number sources appear in Tables~\ref{appendix:tab:lra-full} and~\ref{appendix:tab:wikitext-full}; the WikiText-103 setup and window-size controls are described in Appendices~\ref{appendix:wikitext} and~\ref{sec:window-size}.

\section{Discussion}

We approached ICL in LRNNs as a question of inductive bias: what recurrent update lets a fixed forward pass fit a predictor to its in-context examples?
GRIL answers with a short-window cross-product update that accumulates the full-context gradient statistic and applies the resulting mini-batch step at a multiplicative readout, keeping the $\mathcal{O}(f^2)$ degrees of freedom of the regression problem rather than the $\mathcal{O}(f^4)$ of an indirect gated-recurrent realization of linear attention.
On the synthetic tasks trained GRILs recover the predicted outputs and, for regression, the predicted parameters, so exposing the right local cross-products equips an LRNN with a gradient-descent-style inductive bias for in-context learning without full quadratic attention.

\paragraph{Limitations.}
The exact GD equivalence depends on structured tokenizations that place the input, target, and query factors in each local window.
The non-linear extension is only approximate, since the linear sufficient statistics in the state cannot in general recover the exact gradient of a non-linear predictor.
Beyond the synthetic setting GRIL does not yet match the strongest sequence models---under a matched WikiText-103 protocol it trails Mamba-2 and Gated DeltaNet-2, and our language-modeling study covers a single dataset and scale.
Characterizing when local cross-product recurrences stay aligned with GD at scale is left to future work.

\subsubsection*{Acknowledgments}
The authors would like to thank Mark Schöne for his insightful feedback and constructive suggestions that greatly improved this work. NMS was funded by the German Federal Ministry of Research, Technology and Space (BMFTR) project SMILES (01GQ2504A). YT and DK were funded by project SAIL (grant no. NW21-059A). The authors gratefully acknowledge the Gauss Centre for Supercomputing e.V. (\url{www.gauss-centre.eu}) for funding this project by providing computing time on the GCS Supercomputer JUWELS at Jülich Supercomputing Centre (JSC).
The authors also acknowledge the ZIH at TU Dresden for providing additional HPC resources that supported this research.

\bibliography{references}
\bibliographystyle{apalike}

\clearpage
\appendix

\section{Theoretical derivations}

\subsection{Scalar linear regression} \label{appendix:1dlinreg}

As linear regression is performed on the training dataset $\mathcal{D} = \{\langle \vx_i, y_i \rangle\}_{i=1}^{N}$, the SSM receives the training data in the form of a sequence as input. 
In the most general case, this is a sequence $\vs_1 = \vx_1, \vs_2 = [0, \ldots, y_1], \ldots$, where $[\ldots]$ denotes concatenation and $y_i$ is padded with $f-1$ zeros for its dimensions to match that of $\vx_i$.
This more general case is discussed in the next section.
But here, we will consider a case which simplifies our construction.

Let $\vs_1, \vs_2, \ldots$ be a sequence of constructed context vectors with value and query embeddings $\vv_t = \vq_t = \vs_t = [\vx_t y_t,\vx_{t+1}] \in \R^{2f}$, and let us assume $\vw_0 = \vzero$ for simplicity \footnote{The more general case is treated for the case of multi-step GD in Appendix~\ref{appendix:multi}.}.
If the sequence input map $\mPsi \in \R^{f \times 2f}$ is such that $\mPsi \vv_t = \vx_t y_t$, Eq.~\ref{eq:1dlrnn} can be written as an SSM (Eq.~\ref{eq:ssm}), i.e.
\begin{align}
    \vz_t &= \mI \, \vz_{t-1} + \mPsi \, \vv_t \label{eq:1dssm}\,.
\end{align}
A simple choice is $\mPsi=[\mI_f\;\vzero]$, which selects the first $f$ entries of $\vv_t$.
The state of this network is the negative unscaled gradient, $\vz_t = -t\,\nabla_{\vw} \L(\D_{1:t}; \vw_0)$, and the state recursion accumulates these terms.

The accumulated gradient is then `applied' to the implicit model's initial parameters, $\vw_0$, before computing the $(N+1)$-th output.
With $\vw_0 = \vzero$, the output is 
\begin{align}
    o_t &= \beta \vz_t^T \mTheta \vq_t  \,, \label{eq:1dssmout}
\end{align}
where $\beta=\frac{\eta}{t}$, $\eta$ is the learning rate, and $t$ is the number of context examples seen so far ($t=N$ at the test query).
The SSM final output, $o_t$ above, corresponds to a prediction of the trained linear model, $\hat{y}_{t+1}$ in Eq.~\ref{eq:predy}, if the query selector $\mTheta\in\R^{f\times 2f}$ obeys $\mTheta \vq_t = \vx_\tpone$.
For example, $\mTheta=[\vzero\;\mI_f]$ selects the last $f$ entries of $\vq_t$ (see Figure~\ref{fig:lossnmatrices} for a concrete example).
Note that the output in  Eq.~\ref{eq:1dssmout} matches the general form of the SSM output in Eq.~\ref{eq:ssmout} (without the non-linearity).

The above shows that the following SSM
\begin{equation}
\label{eq:overall}
\begin{aligned}
    \vv_t = \vq_t &= [\vx_t y_t, \vx_\tpone]\,, \\
    \vz_t &= \mI \, \vz_{t-1} + \mPsi \, \vv_t\,,  \\
    o_t &= \beta \vz_t^T \mTheta \vq_t = \hy_\tpone\,.
\end{aligned}
\end{equation}
can perform gradient descent on the parameters $\vw_0$ of the implicit linear model and use this mechanism for in-context learning.
The specific structure of the SSM in \eqref{eq:overall} demonstrates the importance of multiplicative processing, for both the inputs and outputs.

\subsection{Multivariate linear regression} \label{appendix:ndlinreg}

The implicit linear model for multivariate linear regression is 
\begin{equation*}
    \hvy = \mW^T \vx \,,
\end{equation*}
for $\mW \in \R^{f \times f}$ and $\hvy, \vx \in \R^f$.
The loss is
\begin{align}
    \L(\D; \mW_0)  = \frac{1}{2N} \sum_i \left(\mW_0^T \vx_i - \vy_i \right)^T\,\left(\mW_0^T \vx_i - \vy_i \right) \,. \label{eq:appendix-ndloss}
\end{align}

The gradient of this loss over the first $t$ samples $\D_{1:t}$ is
\begin{align}
\nabla_{\mW}\L(\D_{1:t};\mW_0)
&=
\left.
\frac{1}{2t}\sum_{i=1}^{t}
\nabla_{\mW}
\left\|
\mW^\top\vx_i-\vy_i
\right\|_2^2
\right|_{\mW=\mW_0}
=
\frac{1}{t}\sum_{i=1}^{t}
\vx_i
\left(
\mW_0^\top\vx_i-\vy_i
\right)^\top .
\label{eq:gradient}
\end{align}

For notational simplicity, we will write this as $f$ scalar regression problems, one for each element of $\vy$.
Using $[\mW]_{:,i}$ to denote the $i$-th column of matrix $\mW$, using $[\mW_0]_{:,i}$ to denote the $i$-th column of matrix $\mW_0$ and using $[\vy_t]_i$ to denote the $i$-th element of the target vector $\vy_t$,
\begin{align}
\vg(\D_{1:t};[\mW_0]_{:,i})
&= \vg(\D_{1:t-1};[\mW_0]_{:,i})
+ \left([\mW_0]_{:,i}^{T}\vx_t-[\vy_t]_i\right)\vx_t .
\label{eq:ndgrad}
\end{align}

where $\frac{1}{t} \vg(\D_{1:t}; [\mW_0]_{:,i}) = \left[\nabla_{\mW} \L(\D_{1:t}; \mW_0)\right]_{:,i}$.

\paragraph{Implementation as an SSM:}

The input is given using the value and key embeddings constructed as in Eq.~\ref{eq:context}.

This calculates $\mV_t \mC \mK_t^T$, which is then provided as input to the SSM layer.
If $\mC = \left(\begin{smallmatrix}0 \\ 1 \\ 0\end{smallmatrix}\right) \left(\begin{smallmatrix}1 \\ 0 \\ 0\end{smallmatrix}\right)^T = \left[\begin{smallmatrix}0 & 0 & 0 \\ 1 & 0 & 0 \\ 0 & 0 & 0\end{smallmatrix}\right]$, this corresponds to the input to the SSM being $\vy_t\vx_t^T$ (assuming $\mW_0=\vzero$).

Defining $\mZ_t \in \R^{f \times f}$ as the state of the SSM, where $[\mZ_t]_{i, :} = -\vg(\D_{1:t}; [\mW_0]_{:,i})^T$, equivalently $\mZ_t = -t\left(\nabla_{\mW}\L(\D_{1:t};\mW_0)\right)^T$, and assuming $\mW_0 = \vzero$, the equations Eq.~\ref{eq:ndgrad} for all $i$ (all rows of $\mZ_t$) can be written as a single equation
\begin{align*}
    \mZ_t =  \mZ_{t-1} + \mV_t \mC \mK_t^T \,,
\end{align*}
with output
\begin{align*}
    \vo_t =  \frac{\eta}{t} \mZ_t \, \mQ_t \, \vc \,.
\end{align*}
If $\vc = \left(\begin{smallmatrix} 0 \\ 0 \\ 1 \end{smallmatrix}\right)$, this corresponds exactly to the output $\hvy_\tpone = -\eta \nabla_{\mW} \L(\D_{1:t}; \mW_0)^T\,\vx_\tpone$.
This provides a construction that allows us to perform gradient descent on the parameters of an arbitrary dimensional input model.

\subsection{Multi-step GD} \label{appendix:multi}

\begin{proposition}
    Given $1, \ldots, L$ diagonal linear recurrent layers that receive input from a sliding window of size 3 and stride 2, and sequence tokens $\vs_{2j-1} = \vx_j$ and $\vs_{2j} = \vy_j$, for $j = 1, \ldots, N$, drawn from a linear model, for each layer $l$, one can construct pairs of recurrent matrices $\mA^{(l,1)},\mA^{(l,2)}$, inputs $\mB^{(l,1)},\mB^{(l,2)}$ and output matrices $\mU^{(l,1)}, \mU^{(l,2)}$ such that each recurrent step $j$ produces $\hat{\vy}_{j+1} = \mW_l^T \vx_{j+1}$ as output, where $\mW_l = \mW_{l-1} - \eta \nabla_\mW \L(\D_{1:j}; \mW_{l-1})$ is the $l$-th step of gradient descent, started from $\mW_0 = \vzero$.
    The test input $\vx_{N+1}$ is contained in token $\vs_{2N+1}$, and produces the test prediction $\hat{\vy}_{N+1}$.
\end{proposition}

We have so far assumed that $\mW_0 = \vzero$, which for the first layer corresponds to initializing the parameters of the equivalent linear model to all zeros.
For multi-step GD, the layers from the second onwards have a non-zero initial value of parameters, so we derive a general form for a layer that performs GD with non-zero initialization.

Starting from Eq.~\ref{eq:ndgrad}, repeated below for convenience,
\begin{align*}
    \vg(\D_{1:t}; [\mW_0]_{:,i}) &= \vg(\D_{1:t-1}; [\mW_0]_{:,i}) + \left( [\mW_0]_{:,i}^T \, \vx_t - [\vy_t]_i \right) \vx_t \,, 
\end{align*}
we note that this accumulates two different components -- $\left([\mW_0]_{:,i}^T\, \vx_t \, \right) \vx_t$ and $[\vy_t]_i \, \vx_t$.

We propose having two different heads (per layer) to accumulate these quantities separately (i) $\vy_t \, \vx_t^T$ and (ii) $\vx_t \, \vx_t^T$.

Defining $\mZ_t \in \R^{f \times f}$ as in Appendix~\ref{appendix:ndlinreg}, so that $[\mZ_t]_{i, :} = -\vg(\D_{1:t}; [\mW_0]_{:,i})^T$, the recurrent network corresponding to the accumulation of (i) is the same as before:
\begin{align*}
    \mZ_t =  \mZ_{t-1} + \mV_t \mC \mK_t^T \,.
\end{align*}
For (ii), we can write an equivalent recurrent network
\begin{align*}
    \tilde{\mZ}_t =  \tilde{\mZ}_{t-1} + \mV_t \tilde{\mC} \mK_t^T \,,
\end{align*}
If $\tilde{\mC} = \left(\begin{smallmatrix}1 \\ 0 \\ 0\end{smallmatrix}\right) \left(\begin{smallmatrix}1 \\ 0 \\ 0\end{smallmatrix}\right)^T = \left[\begin{smallmatrix}1 & 0 & 0 \\ 0 & 0 & 0 \\ 0 & 0 & 0\end{smallmatrix}\right]$, this corresponds to the input to the SSM being $\vx_t \vx_t^T$.

The output at layer $l$ is
\begin{align*}
    \vo_t^{(l)} = \left( \mW_{l-1}+ \Delta_l \mW \right)^T \, \mQ_t \, \vc \,,
\end{align*}
where
\begin{align*}
    \Delta_l \mW = -\frac{\eta}{t} \left(\tilde{\mZ}_t^{(l)}\mW_{l-1}-\left(\mZ_t^{(l)}\right)^T\right)\,
\end{align*}

In summary, at each layer, there are two linear recurrent layers, and the output includes a multiplicative combination across layers and with the external input.

Since each layer performs the same operation, the construction can also be realized by weight tying: looping a single GRIL block $L$ times gives $L$ gradient steps without adding parameters, so the number of steps is not bounded by depth.
Note further that neither accumulator depends on the layer index, since $\mZ_t=\sum_{i\le t}\vy_i\vx_i^T$ and $\tilde{\mZ}_t=\sum_{i\le t}\vx_i\vx_i^T$ are functions of the data alone; the parameters $\mW_{l-1}$ enter only multiplicatively at the readout.
A looped implementation therefore computes the two sequence scans once, at cost $\mathcal{O}(Tf^2)$, after which each additional gradient step is the recursion
\begin{align*}
    \mW_l = \mW_{l-1} - \frac{\eta}{t}\left(\tilde{\mZ}_t\mW_{l-1}-\mZ_t^T\right)\,.
\end{align*}
Forming $\mW_l$ explicitly costs $\mathcal{O}(f^3)$ per step, since $\tilde{\mZ}_t\mW_{l-1}$ multiplies two dense $f\times f$ matrices, but the prediction never requires it.
Writing $\beta = \eta/t$ and $\mA = \mI - \beta\tilde{\mZ}_t$, the recursion unrolls from $\mW_0 = \vzero$ to $\mW_L = \beta\sum_{k=0}^{L-1}\mA^k\mZ_t^T$, and $\tilde{\mZ}_t = \sum_{i\le t}\vx_i\vx_i^T$ is symmetric, so the readout is
\begin{align*}
    \hat{\vy}_{t+1} = \mW_L^T\vx_{t+1} = \beta\,\mZ_t\sum_{k=0}^{L-1}\mA^k\vx_{t+1}\,.
\end{align*}
Propagating the query through the loop, $\vv_0 = \vx_{t+1}$ and $\vv_{k+1} = \vv_k - \beta\tilde{\mZ}_t\vv_k$, makes every iteration a matrix-vector product, so each additional gradient step costs $\mathcal{O}(f^2)$ and is independent of the sequence length.
The parameter-space form is the one to use when $\mW_l$ itself is needed, as in the parameter comparisons of Section~\ref{sec:exp}.
We do not evaluate this looped variant here and leave it to future work.

\subsection{Depth, looping, and the number of gradient steps} \label{appendix:looping}

A multilayer GRIL can be written as a looped model. From Appendix~\ref{appendix:multi}, the predictor after layer $l$ satisfies

\[
\mW_l
=
\mW_{l-1}
-
\tfrac{\eta}{t}
\bigl(
\tilde{\mZ}_t \mW_{l-1}
-
\mZ_t^T
\bigr).
\]

where

\[
\mZ_t=\sum\nolimits_{i\le t}\vy_i\vx_i^T,
\qquad
\tilde{\mZ}_t=\sum\nolimits_{i\le t}\vx_i\vx_i^T.
\]

Both accumulated states depend only on the input sequence and are unchanged across layers. Defining

\[
F(\mW)
=
\mW
-
\tfrac{\eta}{t}
\bigl(
\tilde{\mZ}_t\mW-\mZ_t^T
\bigr).
\]

the multilayer construction is therefore

\[
    \mW_l = F(\mW_{l-1}),
\]

so $L$ GRIL layers can be replaced by $L$ applications of one weight-tied GRIL block.

This is a restricted form of looping. The original sequence and its local windows remain fixed across iterations, while only the quantity required for the next gradient step is carried between iterations. A general looped layer may feed its complete output representation into the next iteration; the GRIL construction uses a fixed feedback structure that preserves the original context and propagates the current task update.

This places the multi-step construction in the same broad class as looped models that repeatedly apply a shared transformation~\citep{gatmiry_looped_2024,gatmiry_depth_2024,chen_bypassing_2024}. The role of the loop is different from fixed-point recurrence~\citep{movahedi_fixedpoint_2025} or iterative memory refinement~\citep{deng_lt2_2026}: here each iteration corresponds to another gradient step on the same in-context objective.

The recurrent statistics need to be computed only once. Their sequence scan costs $\mathcal{O}(Tf^2)$. If only the prediction is required, the loop can propagate the query through

\[
\vv_{k+1}
=
\vv_k
-
\tfrac{\eta}{t}\tilde{\mZ}_t\vv_k.
\]

with $\vv_0=\vx_{t+1}$, so that $\vv_k=\mA^k\vx_{t+1}$ and the $L$-step prediction is recovered by the final accumulation $\hat{\vy}_{t+1}=\beta\,\mZ_t\sum_{k=0}^{L-1}\vv_k$.
Every additional gradient step therefore requires an $\mathcal{O}(f^2)$ matrix--vector product and is independent of the sequence length. Explicitly forming $\mW_l$ instead requires an $\mathcal{O}(f^3)$ matrix--matrix product per step.

\subsection{Non-linear GD} \label{appendix:nonlinear}

\begin{proposition}
    Given a diagonal linear recurrent layer that receives input from a sliding window of size 3 and stride 2, followed by an MLP layer $\rho$, and tokens $\vs_{2j-1} = \vx_j$ and $\vs_{2j} = y_j$, for $j = 1, \ldots, N$, drawn from a non-linear model, one can construct recurrent matrix $\mA$, input $\mB$ and output matrix $\mU$ such that, for $\vw_0 = \vzero$, each recurrent step $j$ produces $\hat{y}_{j+1} \approx \vw_1^T \vx_{j+1}$ as output, where $\vw_1 = \vw_0 - \eta \nabla_{\vw} \L(\D_{1:j}; \vw_0)$ is one step of gradient descent on the non-linear objective.
    Composing this readout with the output non-linearity of the block gives the prediction $h(\vw_1^T \vx_{j+1})$ of the updated model.
    The equality is approximate because $\rho$ must map the linearly accumulated state onto the non-linear gradient, which cannot in general be represented exactly.
    The test input $\vx_{N+1}$ is contained in token $\vs_{2N+1}$, and produces the test prediction $\hat{y}_{N+1}$.
\end{proposition}

Let us consider a non-linear regression problem with a least squares loss, where the outputs are scalar for simplicity.
For a given dataset of $N$ samples $\mathcal{D} = \{\langle \vx_i, y_i \rangle\}_{i=1}^{N}, \vx \in \R^f, y \in \R$, 
predictions from a non-linear model are generated using
\begin{equation*}
    \hat{y} = h(\vw^T \vx) \,, 
\end{equation*}
for $\vw \in \R^{f}$ and where $h$ is some non-linear function such as sigmoid or an MLP.
This will be our implicit non-linear model for the scalar-target case.

The best fit for $\vw$ is found by minimizing the loss
\begin{align*}
    \L(\D; \vw_0) &= \frac{1}{2N} \sum_{i=1}^N || \hat{y}_i - y_i ||_2^2 \\
    &= \frac{1}{2N} \sum_i \left(h(\vw_0^T \vx_i) - y_i \right)^2 \,. 
\end{align*}

The gradient of the loss calculated on the first $t$ samples of the dataset is
\begin{align*}
    \nabla_{\vw} \L(\D_{1:t}; \vw_0) &= \frac{1}{t} \sum_{i=1}^t \left( h(\vw_0^T \, \vx_i) - y_i \right) h'(\vw_0^T \, \vx_i) \vx_i \,,
\end{align*}
where $\D_{1:t}$ denotes the first $t$ samples in $\D$, and $h'$ denotes the first derivative of $h$. 

For comparison with the linear recurrence, define the auxiliary accumulated vector
\begin{align*}
    \vg_{\vw_0} (\D_{1:t}) &= \sum_{i=1}^t \left( y_i - \vw_0^T \, \vx_i \right) \vx_i \,,
\end{align*}
which is the quantity the linear construction accumulates and can be recursively calculated as before.
For a non-linear $h$ it no longer equals the negative unscaled gradient.

Writing this as an SSM exactly as in \ref{eq:1dssm} (with the same $\mPsi$)
\begin{align*}
    \vz_t &= \mI \, \vz_{t-1} + \mPsi \, \vv_t \,,
\end{align*}
we now need to calculate the output as
\begin{align*}
    \hat{y}_{t+1} &=  - \eta \nabla_{\vw_0} \L(\D_{1:t}; \vw_0)^T \, \vx_{t+1} \,,
\end{align*}
taking $\vw_0 = \vzero$ as in Appendix~\ref{appendix:ndlinreg}, so that this expression is $\vw_1^T \vx_{t+1}$.

The interleaved MLP layer $\rho$ would need to learn a mapping such that
\begin{align*}
    \rho\left(\sum_{i=1}^t\left( y_i - \vw_0^T \, \vx_i \right) \vx_i \right) &\approx  \sum_{i=1}^t\left( y_i - h(\vw_0^T \, \vx_i) \right) \times h'(\vw_0^T \, \vx_i) \vx_i\,
\end{align*}
which cannot in general be represented exactly due to the information compression induced by the linear accumulation.
Concretely, at $\vw_0=\vzero$ the linear state retains only $\sum_i y_i \vx_i$, whereas the exact gradient is $\frac{1}{t}\sum_i \left(h(0)-y_i\right)h'(0)\,\vx_i = \frac{h'(0)}{t}\left(h(0)\sum_i \vx_i - \sum_i y_i\vx_i\right)$, which additionally requires the statistic $\sum_i \vx_i$ whenever $h(0)\neq 0$ (for example $h(0)=\tfrac{1}{2}$ for a sigmoid).
Since $\vz_t$ does not retain $\sum_i \vx_i$, the mapping cannot be exact in general.
Instead, a sufficiently expressive MLP can learn an approximate mapping that captures this non-linear transformation over the training distribution.
This can then be used to calculate the output as
\begin{align*}
    o_t &= \frac{\eta}{t}\rho(\vz_t)^T \mTheta \vq_t   \,.
\end{align*}
The readout is therefore the linear response $\vw_1^T \vx_{t+1}$ of the updated parameters, and not yet the prediction of the updated non-linear model; the outer non-linearity is supplied by composing it with the output map in Eq.~\ref{eq:ssmout}.

\subsection{Regularization terms in the loss} \label{appendix:regularisation}

A penalty term that depends only on the current parameters, and not on the context, can be folded into the same readout, because its gradient is available without any additional recurrent statistic.
For example, adding an L2 regularization term to the loss in Eq.~\ref{eq:appendix-ndloss} gives
\begin{align*}
    \L(\D_{1:t}; \mW) = \frac{1}{2t} \sum_{i=1}^t || \hat{\vy}_i - \vy_i ||_2^2 + \frac{\lambda}{2t} || \mW ||_F^2 \,,
\end{align*}
where the penalty carries the same $1/t$ factor as the data term, so that $\lambda$ fixes the ratio between the two and the whole gradient is read out with the single coefficient $\eta/t$.
Against a standard ridge penalty $\frac{\lambda'}{2}||\mW||_F^2$, this corresponds to $\lambda' = \lambda/t$, that is, a regularizer normalized by context length; a context-independent $\lambda'$ is implemented in the same way, by leaving its term unscaled in the update below.
The gradient on the first $t$ samples is
\begin{align*}
    \nabla_{\mW} \L(\D_{1:t}; \mW_0) = \frac{1}{t} \sum_{i=1}^t \vx_i \left(\mW_0^T \vx_i - \vy_i \right)^T + \frac{\lambda}{t} \mW_0 \,.
\end{align*}
To construct an SSM that does GD on this loss, we change the $\Delta_l \mW$ used in Appendix~\ref{appendix:multi} to
\begin{align*}
    \Delta_l \mW = -\frac{\eta}{t} \left(\tilde{\mZ}_t^{(l)}\mW_{l-1}-\left(\mZ_t^{(l)}\right)^T + \lambda \mW_{l-1} \right)\,.
\end{align*}

\subsection{Classification losses}

\label{appendix:celoss}

We now show that the gradient-generation structure used in the linear-regression construction also appears in standard classification losses, for both the softmax and the binary case.

\paragraph{Softmax cross-entropy.}
Consider a $K$-class linear classifier with input $\vx_i\in\mathbb{R}^f$, feature map $\phi(\vx_i) \in \mathbb{R}^d$, and weight matrix $\mW\in\mathbb{R}^{K\times d}$, matching the convention of Section~\ref{sec:nonlinear}.%
The logits and predicted class probabilities are
\[
    \vz_i = \mW \phi(\vx_i) \in \mathbb{R}^K,
    \qquad
    \vp_i = \mathrm{softmax}(\vz_i),
\]
where
\[
    [\vp_i]_k
    =
    \frac{\exp([\vz_i]_k)}
    {\sum_{j=1}^{K}\exp([\vz_i]_j)} .
\]
For a one-hot label vector $\vy_i\in\{0,1\}^K$, the single-sample cross-entropy loss is
\[
    \ell_{\mathrm{CE}}(\mW;\vx_i,\vy_i)
    =
    -\sum_{k=1}^{K}[\vy_i]_k \log [\vp_i]_k .
\]

Using the standard softmax identity,
\[
    \frac{\partial [\vp_i]_k}{\partial [\vz_i]_j}
    =
    [\vp_i]_k
    \left(
        \delta_{kj} - [\vp_i]_j
    \right),
\]
the gradient of the loss with respect to the logits is
\[
    \nabla_{\vz_i}
    \ell_{\mathrm{CE}}(\mW;\vx_i,\vy_i)
    =
    \vp_i - \vy_i .
\]
Since $[\vz_i]_k = [\mW]_{k,:}\,\phi(\vx_i)$, the gradient with respect to the $k$-th row of $\mW$ is
\[
    \frac{\partial \ell_{\mathrm{CE}}}{\partial [\mW]_{k,:}}
    =
    \bigl([\vp_i]_k - [\vy_i]_k\bigr)\phi(\vx_i)^T .
\]
Stacking these row-wise gradients gives
\[
    \nabla_{\mW}
    \ell_{\mathrm{CE}}(\mW;\vx_i,\vy_i)
    =
    (\vp_i-\vy_i)\phi(\vx_i)^T
    \in \mathbb{R}^{K\times d}.
\]
Thus, the per-sample classification gradient has the same outer-product form as in the regression case, with the scalar residual replaced by the vector-valued prediction error $\vp_i-\vy_i$.

For a dataset $\mathcal{D}_{1:N}=\{(\vx_i,\vy_i)\}_{i=1}^N$, the total gradient is therefore
\[
    \nabla_{\mW}
    \mathcal{L}_{\mathrm{CE}}(\mathcal{D}_{1:N};\mW)
    =
    \sum_{i=1}^{N}
    (\vp_i-\vy_i)\phi(\vx_i)^T .
\]
Equivalently, for the first $t$ samples, this gradient can be accumulated recursively as
\begin{align*}
    \nabla_{\mW}\mathcal{L}_{\mathrm{CE}}(\mathcal{D}_{1:t};\mW)= \nabla_{\mW}\mathcal{L}_{\mathrm{CE}}(\mathcal{D}_{1:t-1};\mW)+ (\vp_t-\vy_t)\phi(\vx_t)^T ,
\end{align*}
where $\vp_t=\mathrm{softmax}(\mW\phi(\vx_t))$.

\paragraph{Binary cross-entropy.}\label{appendix:sigloss}
Binary classification can be viewed as the single-logit analogue of the softmax case. 
Let $\vw\in\mathbb{R}^d$ be the classifier weight vector acting on the same feature map, and define
\[
    z_i = \vw^T \phi(\vx_i),
    \qquad
    p_i = \sigma(z_i)
    =
    \frac{1}{1+\exp(-z_i)} ,
\]
with label $y_i\in\{0,1\}$. 
The single-sample binary cross-entropy loss is
\[
    \ell_{\mathrm{BCE}}(\vw;\vx_i,y_i)
    =
    -
    \left[
        y_i\log p_i
        +
        (1-y_i)\log(1-p_i)
    \right].
\]
Using $\frac{\partial p_i}{\partial z_i}=p_i(1-p_i)$, the derivative with respect to the logit is
\[
    \frac{\partial \ell_{\mathrm{BCE}}}{\partial z_i}
    =
    p_i-y_i .
\]
Since $z_i=\vw^T\phi(\vx_i)$, the gradient with respect to $\vw$ is
\[
    \nabla_{\vw}
    \ell_{\mathrm{BCE}}(\vw;\vx_i,y_i)
    =
    (p_i-y_i)\phi(\vx_i) .
\]
For the first $N$ samples, the total gradient is
\[
    \nabla_{\vw}
    \mathcal{L}_{\mathrm{BCE}}(\mathcal{D}_{1:N};\vw)
    =
    \sum_{i=1}^{N}
    (p_i-y_i)\phi(\vx_i) .
\]
The corresponding recursive accumulation is
\begin{align*}
    \nabla_{\vw}\mathcal{L}_{\mathrm{BCE}}(\mathcal{D}_{1:t};\vw)= \nabla_{\vw}\mathcal{L}_{\mathrm{BCE}}(\mathcal{D}_{1:t-1};\vw) + (p_t-y_t)\phi(\vx_t) ,
\end{align*}
where $p_t=\sigma(\vw^T\phi(\vx_t))$.

Both losses therefore share the accumulation structure of the regression setting, with the error component given by the difference between the model prediction and the target label.
For softmax cross-entropy the prediction error is vector-valued, $\vp_i-\vy_i$, and the accumulated gradient is a matrix in $\mathbb{R}^{K\times d}$, while for binary cross-entropy it is scalar, $p_i-y_i$, and the accumulated gradient is a vector in $\mathbb{R}^{d}$.
In both cases the local terms can be accumulated recurrently and then used by the readout to implement a gradient-like update, which is the structural property GRIL relies on for regression. The GRIL value embedding $\mV_t$ then holds the error vector $\vy_t-\vp_t$, the key embedding $\mK_t$ holds $\phi(\vx_t)$, and the query embedding $\mQ_t$ holds $\phi(\vx_{t+1})$, which requires padding or projecting them into a common ambient token dimension, since the error vector lies in $\mathbb{R}^{K}$ while the features lie in $\mathbb{R}^{d}$.
Forming $\vy_t-\vp_t$ directly from the sequence tokens is exact only under the construction's $\mW_0=\vzero$ assumption: there $\vp_t$ is the constant uniform prediction (for example $1/K$ for softmax), so the residual is a fixed affine shift of the observed label $\vy_t$.
For $\mW_0\neq\vzero$, $\vp_t$ depends non-linearly on $\vx_t$ and would have to be computed by a preceding layer rather than read directly from the input.

\section{Synthetic in-context regression}

This section provides the experimental details and ablations for the synthetic regression tasks reported in the main text. The ablations remove the input construction and multiplicative readout used by the analytic GRIL construction. All runs used a single NVIDIA A100 GPU and took at most 4 hours.

\subsection{Scalar regression} \label{appendix:expt-1d-regression}

The dataset construction and the model evaluation methods are the same as those used in~\citep{von_oswald_transformers_2023}.
Each task (context) $\tau$ consists of in-context training data $D_{\tau} = \left\{ (x_{\tau,i}, y_{\tau,i}) \right\}_{i=1}^{N}$ and test point $(x_{\tau,N+1}, y_{\tau,N+1})$.
At every optimization step of the model, we sample the regression parameters  $W_{\tau} \sim \mathcal{N}(0, I)$. 
We then sample $x_{\tau,i} \sim U(-1, 1)^{f}$ and construct a scalar target $y_{\tau,i} = W_{\tau} x_{\tau,i}$, where ${f}$ is the feature size of the input. 
To evaluate the model, a set of $10^4$ tasks is sampled and mean squared error is calculated.

We use context length $N=10$ and a single-layer GRIL with hidden dimension 20. No activation function was used during training. We use two metrics to compare the trained GRIL model with the GD. 
The $L_2$ distance measures the magnitude of their difference, i.e.,
\[
\left\|\hat{\vy}_{\theta}(\vx_{\tau,\mathrm{test}})
-
\hat{\vy}_{\mathrm{GD}}(\vx_{\tau,\mathrm{test}})\right\|_2,
\]
and is used to compare either predictions or input sensitivities. 
The cosine similarity measures whether two sensitivity vectors point in the same direction,
\[
\mathrm{cos}(\va,\vb)=\frac{\langle \va,\vb\rangle}{\|\va\|_2\|\vb\|_2},
\]
where $\va=\partial \hat{\vy}_{\theta}/\partial \vx_{\tau,\mathrm{test}}$ and 
$\vb=\partial \hat{\vy}_{\mathrm{GD}}/\partial \vx_{\tau,\mathrm{test}}$. 
Thus, the $L_2$ distance captures absolute discrepancy, while cosine similarity captures directional alignment.

The training process spanned a maximum of 300,000 steps with a batch size of 64. A cosine annealing schedule and linear warmup were utilized for the learning rate, beginning with an initial SSM learning rate of $1 \times 10^{-4}$ for optimizing the SSM parameters using the AdamW optimizer. The global learning rate for the remaining parameters was set to $2 \times 10^{-4}$, and these parameters were also optimized using AdamW. A weight decay of 0.05 was applied.

This experiment analyzes the impact of the scalar counterparts of the two mechanisms studied in the main text, the input construction and the multiplicative output construction.
Although the scalar regression setting does not involve a sliding window, the input construction plays an analogous role to the sliding-window input mechanism in the higher-dimensional setting.
Similarly, the output construction is the scalar analogue of the multiplicative output skip-connection.
The results show that the model achieves the lowest loss when both components are present ($0.209$), whereas removing either component, or removing both, increases the loss to $0.426$.

\begin{table}[tbp]
\centering
\caption{Ablation test on GRIL for scalar linear regression.}
\label{appendix:ablation-1d}

\setlength{\tabcolsep}{3pt}
\begin{tabular}{lccc}
\toprule
Model &
\shortstack{Input\\Construction} &
\shortstack{Output\\Construction} &
Loss \\
\midrule
GRIL & $\surd$  & $\surd$  & $0.209 \pm 0.0012$ \\
GRIL & $\times$ & $\surd$  & $0.426 \pm 0.0016$ \\
GRIL & $\surd$  & $\times$ & $0.426 \pm 0.0016$ \\
GRIL & $\times$ & $\times$ & $0.426 \pm 0.0016$ \\
\bottomrule
\end{tabular}
\end{table}

\subsection{Multivariate regression} \label{appendix:expt-nd-regression}

The setup is similar to scalar regression, but we sample $n_{o}$ different regression parameters $W_{\tau}^{k} \sim \mathcal{N}(0, I)$, where $n_{o}$ is the dimension of the target vector $y_{\tau,i}$ and $1\le k\le n_{o}$. We then construct the target coordinate $y_{\tau,i}^{k} = W_{\tau}^{k} x_{\tau,i}$ for each $k$. 
For all our experiments, we choose $n_{o}=f$. We evaluate the multivariate model on the same quantities as the scalar case, namely training loss, cosine similarity and $L_2$ distance to the one-step GD solution, loss across input dimensions $f$, and loss under input rescaling.

To have the model prediction equivalent to one step gradient descent, we train the single-layer GRIL using the Adam optimizer, with an initial learning rate of 0.0001 for recurrent parameters with cosine annealing. For all the other parameters we double the learning rate that is used for recurrent parameters. In all our experiments, each optimization step contains 64 tasks.

\begin{figure*}[tbp]
    \centering
    \includegraphics[width=\textwidth]{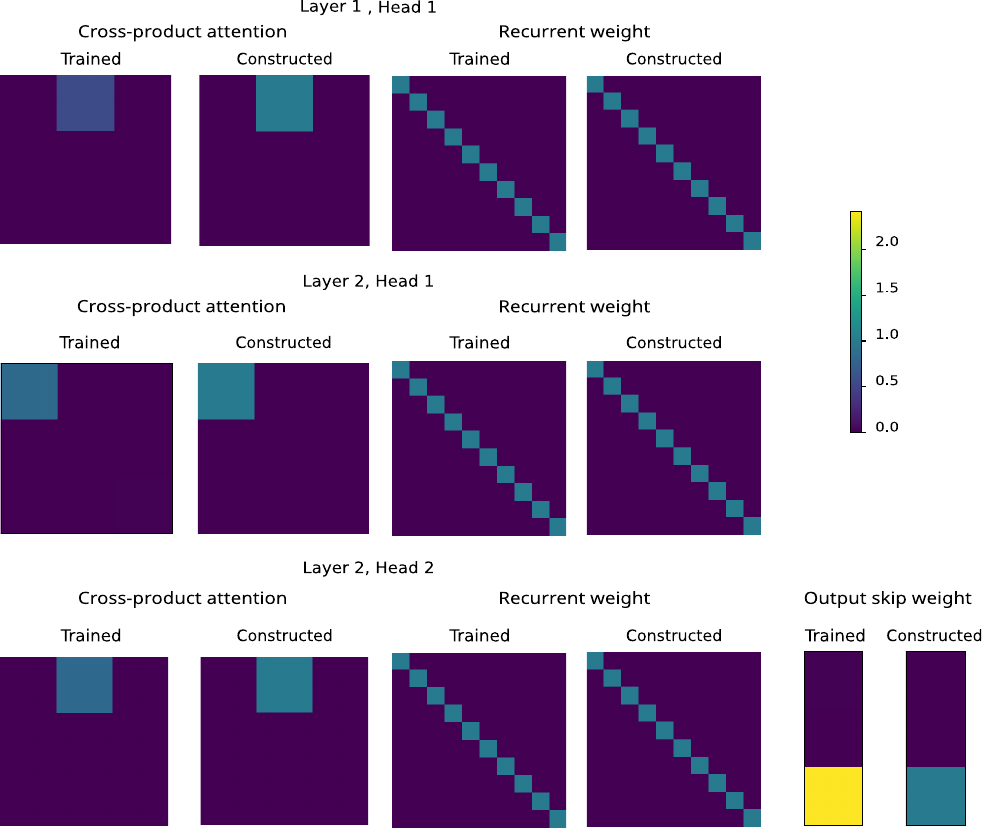}
    \caption{
    Visualization of the trained parameters for a two-layer GRIL on the linear regression task.
    }
    \label{appendix:fig:2layer-weights}
\end{figure*}

The two features predicted by the construction are the sliding-window input, which exposes $(\vx_t,\vy_t,\vx_{t+1})$, and the multiplicative interaction between the recurrent state and the query.
Table~\ref{appendix:ablation_study} compares the full one-layer GRIL with variants that disable the sliding-window input and/or the multiplicative output skip-connection (five seeds).
Both features are required to match one-step GD: removing either approximately doubles the loss ($0.206{\to}0.414$).
This gap is the same order as generic S5 baselines that see the interleaved tokens but lack the cross-product write and multiplicative readout (Table~\ref{appendix:model_comparison}, loss $0.426$).
This comparison isolates the bilinear read-write structure; when identical local tokens are used, the model fails to reproduce the single-step gradient descent (GD) solution.
\begin{table}[tbp]
\centering
\caption{A comparison of GRIL variants on multivariate linear regression, with the sliding-window input and output skip connection turned off in ablations.}
\label{appendix:ablation_study}

\setlength{\tabcolsep}{3pt}
\begin{tabular}{c c c c}
\toprule
Model &
\shortstack{Sliding\\window input} &
\shortstack{Multiplicative \\ output} &
Loss \\
\midrule
GRIL & $\surd$  & $\surd$  & $0.206 \pm 0.001$ \\
GRIL & $\times$ & $\surd$  & $0.414 \pm 0.002$ \\
GRIL & $\surd$  & $\times$ & $0.414 \pm 0.002$ \\
GRIL & $\times$ & $\times$ & $0.414 \pm 0.002$ \\
\bottomrule
\end{tabular}
\end{table}

\input{window_size}
\input{multi_layer_depth}

\subsection{Non-linear regression}
\label{appendix:expt-nonlinear}

For the non-linear regression task, we follow the sine-wave regression task of~\citet{von_oswald_transformers_2023}, with scalar inputs and targets ($f = n_o = 1$). For each task $\tau$ we draw $u_\tau \sim \mathcal{U}[0,1]$, set the amplitude $a_\tau = 0.1 + 0.4\,u_\tau$ and the phase $\varphi_\tau = \pi u_\tau$, and generate $N = 10$ context pairs and one query by
\begin{equation*}
    x_{\tau,i} \sim \mathcal{U}[-2.5,\, 2.5], \quad y_{\tau,i} = a_\tau \sin\left(x_{\tau,i} + \varphi_\tau\right), \quad i = 1, \ldots, N+1,
\end{equation*}
so the frequency is fixed to one and no noise is added. The sequence interleaves $x_{\tau,1}, y_{\tau,1}, \ldots,$ $x_{\tau,N}, y_{\tau,N}, x_{\tau,N+1}$, and the loss is the squared error on $y_{\tau,N+1}$.
On top of the linear GRIL layer, we add a non-linear mapping implemented as a weighted sigmoid gated unit~\citep{TANAKA2020354}, as in~\citep{smith_simplified_2022}.
In this setting, GRIL denotes a fully trained model, while GD$^\dagger$ denotes a model where the recurrent GRIL layers are fixed according to the gradient-descent construction, and only the auxiliary non-linear mapping is trained.
\subsection{Comparison with other sequence models}

The configurations behind Figure~\ref{fig:result-regression-multi-layer}C are as follows.
The GRIL comparison uses a single recurrent layer, whereas the strongest Mamba+Linear and Griffin+Linear variants use four layers.
The base Griffin and Mamba models were trained with Adam at learning rate $0.0001$, matching S5's configuration.
Mamba+Linear used 4 layers and 128 hidden states, trained with AdamW at learning rate $0.0001$, in both the multivariate and the scalar setting.
Griffin+Linear used 4 layers, an LRU width of 128 and 4 Multi-Query Attention (MQA) heads, trained with Adam at learning rate $0.0001$ in the scalar setting and $0.0002$ in the multivariate setting.

\begin{figure*}[tbp]
    \centering
    \includegraphics[width=0.6\textwidth]{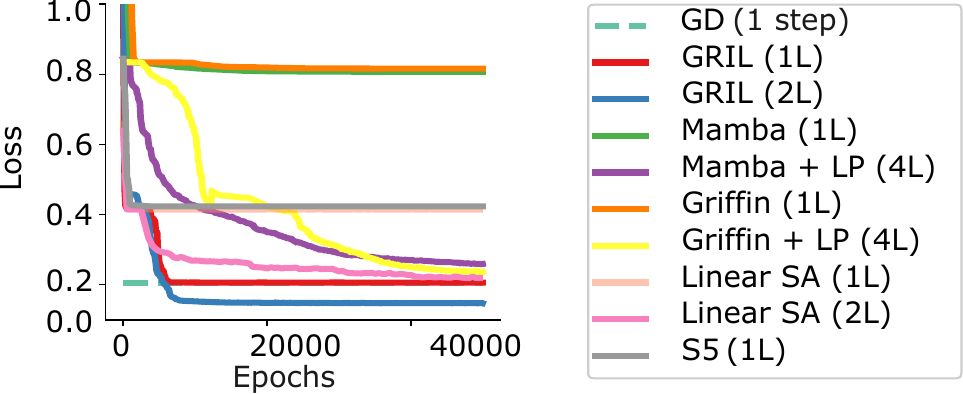}
    \caption{
    Comparison with other models on multivariate linear regression. The GRIL model was evaluated with both 1-layer and 2-layer configurations, and the S5, Mamba(+LP), and Griffin(+LP) models were included for comparison. Linear SA refers to linear self-attention models, with both 1-layer and 2-layer variants tested to evaluate their performance.
    }
    \label{appendix:fig:nd-comparison}
\end{figure*}

\begin{table}[tbp]
\centering
\small
\caption{Comparison of Griffin, Mamba, and S5 models without LP under different numbers of layers and heads for multivariate linear regression.}
\label{appendix:model_comparison}

\setlength{\tabcolsep}{4pt}
\begin{tabular}{lcccc}
\toprule
\multicolumn{5}{c}{\textbf{Griffin Model}} \\
\midrule
\# heads  & 1      & 2      & 2      & 5      \\
\# layers & 1      & 1      & 2      & 1      \\
Loss      & 0.7490 & 0.5665 & 0.4114 & 0.6047 \\
\midrule
\multicolumn{5}{c}{\textbf{Mamba Model}} \\
\midrule
Model dim. & 32     & 64     & 64     & 128    \\
\# layers  & 1      & 1      & 2      & 4      \\
Loss       & 0.5388 & 0.4126 & 0.4179 & 0.4157 \\
\midrule
\multicolumn{5}{c}{\textbf{S5 Model}} \\
\midrule
\# layers & 1     & 2     & 4     & 6     \\
Loss      & 0.426 & 0.426 & 0.426 & 0.426 \\
\bottomrule
\end{tabular}
\end{table}

Table~\ref{appendix:model_comparison} reports a separate layer and head sweep for the same baselines \emph{without} the linear projection, so it measures what those architectures reach on this task on their own.
Griffin combines two RG-LRU modules with one MQA module, while Mamba uses its basic block.
Griffin is best at 2 heads and 2 layers ($0.4114$) and does not improve under further scaling, and Mamba stays between $0.4126$ and $0.5388$ across configurations.
None of these settings reaches the one-step GD loss of $0.207$ (Table~\ref{appendix:second-order-methods}), so the gap is not closed by width or depth alone.

Adding the linear projection changes this: comparing Figure~\ref{appendix:fig:nd-comparison} with Figure~\ref{fig:result-regression-multi-layer}C, Mamba+Linear and Griffin+Linear improve on their base counterparts in Table~\ref{appendix:model_comparison}.

We also sweep the depth of the S5 baseline on the multivariate linear regression task. Each S5 model uses model size and state size $32$ with input and output projections between the data dimension and the model dimension, is trained with Adam at learning rate $0.0001$, and discretizes the state-space system by zero-order hold. Model and state sizes in $\{10, 64, 128, 256\}$ give no improvement over $32$.  

\subsection{Second-order optimization baselines}

\label{lineargd}

To test whether the trained GRIL dynamics are aligned with one step of gradient descent in particular, rather than with any stronger one-step local optimizer, we evaluate one step of the Newton--Raphson method on the evaluation data of Appendix~\ref{appendix:expt-nd-regression} and compare it with trained GRIL and with gradient descent.
Newton--Raphson is a second-order one-step reference that reduces the local objective more aggressively than first-order GD.
The closer agreement with GD is consistent with GRIL recovering a gradient-descent-like mechanism. 

\begin{table}[tbp]
\centering
\small
\caption{Multivariate regression performance of a trained one-layer GRIL model compared with one-step gradient descent (GD), the GD-equivalent GRIL construction, and Newton--Raphson.}
\label{appendix:second-order-methods}

\begin{tabular}{lc}
\toprule
Method & Loss \\
\midrule
Trained GRIL                    & $0.206 \pm 0.001$ \\
Gradient descent (GD)           & $0.207 \pm 0.001$ \\
GRIL (GD-equivalent construction) & $0.206 \pm 0.002$ \\
Newton--Raphson                 & $0.114 \pm 0.001$ \\
\bottomrule
\end{tabular}
\end{table}

\section{In-context classification}

\subsection{Data}

\label{appendix:classificationdata}

Our data generation process is based on an underlying linear regression task, following the multivariate regression framework detailed in Appendix \ref{appendix:expt-nd-regression}. For each sequence, we first sample a weight matrix $\mW \in \mathbb{R}^{f \times K}$ from a standard normal distribution $\mathcal{N}(0, I)$, where $f$ is the input feature dimension and $K$ is the total number of classes. The corresponding input vectors $\vx \in \mathbb{R}^f$ are sampled from a uniform distribution. Subsequently, we compute the logits for each input via a linear transformation: $\vy_{\text{logits}} = \mW^T \vx$.

To adapt this regression setup for classification tasks, we process these logits as follows:

\begin{itemize}
    \item For binary classification ($K=1$), we apply the sigmoid function to the resulting logit and use a threshold of 0.5 to assign the binary label.
    \item For multi-class classification ($K\ge 2$), we transform the logit vector $\vy_{\text{logits}} \in \mathbb{R}^K$ into a probability distribution using the softmax function. The final label is then determined by the index of the maximum logit, which (since softmax is monotonic) equals $\vy_{\text{class}} = \text{argmax}(\vy_{\text{logits}})$.
\end{itemize}

For the classification experiments, an MLP maps the raw input $\vx$ to a non-linear feature representation $\phi(\vx)$.
Unlike the exact construction in Appendix~\ref{appendix:celoss}, the experiments do not form the per-step residual $\vy_t-\vp_t$ explicitly in the local context.
Class probabilities are obtained by applying the output non-linearity after the GRIL readout, while the MLP acts only as an input-side feature map.
The experiments therefore use the same GD$^\dagger$ reference as the non-linear regression setting (Appendix~\ref{appendix:expt-nonlinear}), with the recurrent layers fixed to the construction and only the auxiliary mapping trained.

\subsection{Setup and results}

\label{appendix:clsexp}

We evaluate the single-layer model on several synthetic classification tasks. All results are means over 5 runs with different random seeds, and each run used a single NVIDIA A100 GPU for at most 4 hours.

Learning rate and weight decay were swept at sequence length 10 and the selected values were reused unchanged at sequence length 100, which also tests whether they carry from shorter to longer sequences. Table~\ref{tab:hyperparams} lists the resulting configurations.

\begin{table}[tbp]
\centering
\small
\caption{Optimized hyperparameters from sequence-length-10 experiments.}
\label{tab:hyperparams}

\setlength{\tabcolsep}{3pt}
\begin{tabular}{llcc}
\toprule
\textbf{Dataset} &
\shortstack{\textbf{Input}\\\textbf{Dimension}} &
\shortstack{\textbf{Base Learning}\\\textbf{Rate}} &
\shortstack{\textbf{Weight}\\\textbf{Decay}} \\
\midrule
Sigmoid & 10 & $1.23 \times 10^{-4}$ & $5.17 \times 10^{-3}$ \\
Softmax & 10 & $1.56 \times 10^{-4}$ & $8.45 \times 10^{-5}$ \\
\bottomrule
\end{tabular}
\end{table}

\begin{figure*}[htb]
    \centering
    \includegraphics[width=0.9\textwidth]{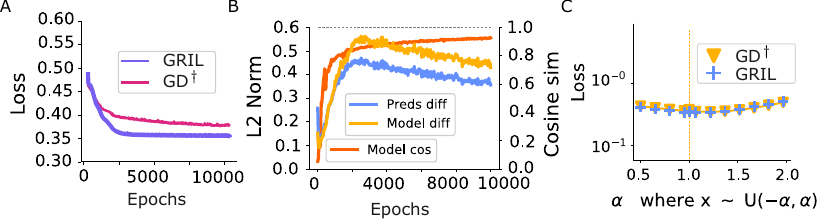}
    \caption{\textbf{Representative classification result.}
    The cross-product construction stays predictive for cross-entropy classification given a learned feature map.
    Single-step GD$^\dagger$ and non-linear GRIL are compared on a 3-class softmax task with sequence length 10.
    GD$^\dagger$ is the non-linear extension of the GD construction (Appendix~\ref{appendix:nonlinear}).
    GRIL tracks GD$^\dagger$ in loss, prediction difference and sensitivity alignment, and remains robust under input rescaling.
    }
    \label{fig:classification-main}
\end{figure*}

\begin{figure*}[htb]
    \centering
    \includegraphics[width=\textwidth]{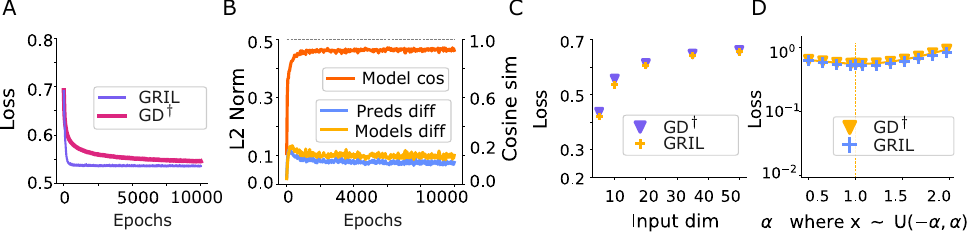}
    \caption{
    \textbf{Comparison of single-step GD$^\dagger$ and non-linear GRIL on a binary classification task with input dimension 10 and sequence length 10.}
    \textbf{A}: Training loss achieved by GD$^\dagger$ and GRIL for the binary-classification case.
    \textbf{B}: Cosine similarity and $L_2$ distance between models and their predictions.
    \textbf{C}: Loss comparison between GD$^\dagger$ and GRIL under varying input dimensions $f$.
    \textbf{D}: Robustness of both models to data scaling (mean and std over 5 seeds shown).}
    \label{fig:appendix:class-sigmoid-10}
\end{figure*}

\begin{figure*}[htb]
    \centering
    \includegraphics[width=0.9\textwidth]{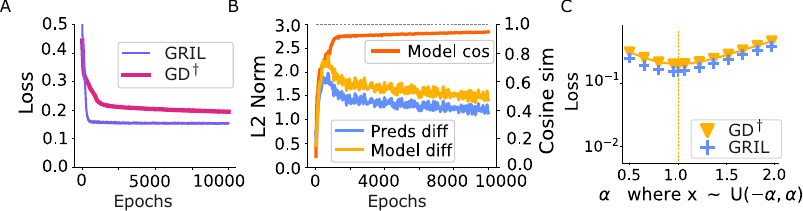}
    \caption{
    \textbf{Comparison of single-step GD$^\dagger$ and non-linear GRIL on a binary classification task with input dimension 10 and sequence length 100.}
    \textbf{A}: Training loss achieved by GD$^\dagger$ and GRIL.
    \textbf{B}: Cosine similarity and $L_2$ distance between models and their predictions.
    \textbf{C}: Robustness of both models to data scaling (mean and std over 5 seeds shown).}
    \label{fig:appendix:class-sigmoid-100}
\end{figure*}

\begin{figure*}[htb]
    \centering
    \includegraphics[width=0.9\textwidth]{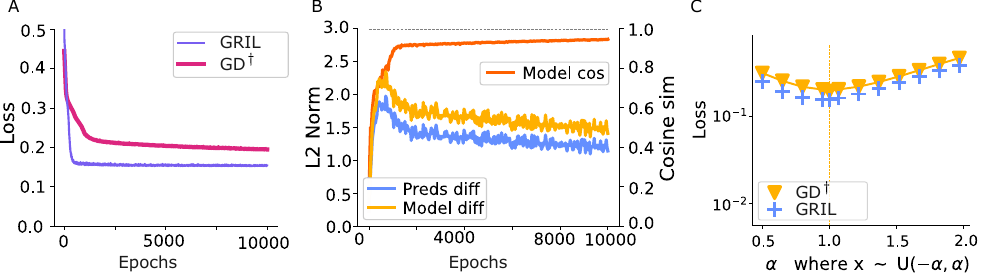}
    \caption{
    \textbf{Comparison of single-step GD$^\dagger$ and non-linear GRIL on a softmax classification task with input dimension 10 and sequence length 100.}
    \textbf{A}: Training loss achieved by GD$^\dagger$ and GRIL.
    \textbf{B}: Cosine similarity and $L_2$ distance between models and their predictions.
    \textbf{C}: Robustness of both models to data scaling (mean and std over 5 seeds shown).}
    \label{fig:appendix:class-softmax-100}
\end{figure*}

Figure~\ref{fig:classification-main} (main text) and Figures~\ref{fig:appendix:class-sigmoid-10}, \ref{fig:appendix:class-sigmoid-100} and~\ref{fig:appendix:class-softmax-100} compare GD$^\dagger$ with GRIL across different classification tasks (binary vs. softmax), input dimensions, and sequence lengths.
GRIL follows the GD$^\dagger$ reference most closely in cross-entropy loss and the direction of the input sensitivity, while prediction-space and parameter-space $L_2$ distances remain non-zero in several settings.
The difference between cosine and $L_2$ agreement is expected because cosine similarity measures direction but not magnitude.
These experiments therefore test approximate functional agreement with GD$^\dagger$ rather than exact recovery of the classification update.

\section{General-purpose sequence tasks}

\subsection{Shared implementation} \label{appendix:general-impl}

The LRA, WikiText-103, and associative-recall experiments use the same GRIL %
blocks as the synthetic experiments, with task-specific input and output projections (Section~\ref{sec:general-tasks}).
The core consists of the local cross-product update, recurrent state accumulation, and multiplicative readout.

\begin{figure}[htb]
    \centering
    \includegraphics[width=0.4\columnwidth]{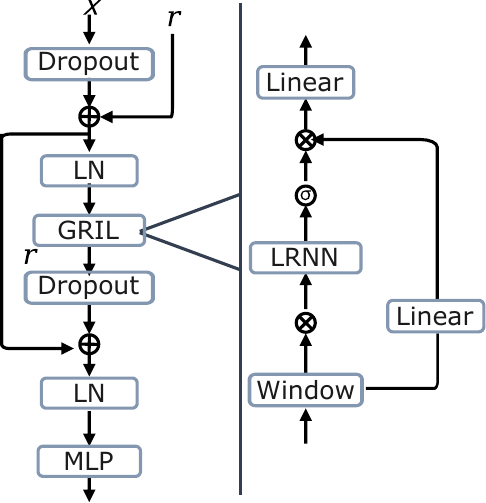}
    \caption{
    Implementation of the GRIL block used in the LRA and WikiText-103 experiments.
    The core GRIL update uses the same windowed cross-product branch as in the
    synthetic experiments, while the full block adds pre-normalization and
    task-level input/output projections.
    }
    \label{appendix:fig:framework}
\end{figure}

As illustrated in Figure~\ref{appendix:fig:framework}, the block adopts a layered structure with pre-norm stabilization. The windowed cross-product branch forms the recurrent input map $\mB=\mV_t\mC\mK_t^T$, where $\mV_t$, $\mK_t$, and $\mQ_t$ are the value, key, and query embeddings of the local window and $\mC$ is the window-mixing matrix. A second branch off the same local window defines the multiplicative readout map $\mU=\mQ_t\vc$, where $\vc$ selects the query column.

\subsection{Long Range Arena}

\subsubsection{Dataset description}

Long Range Arena (LRA)~\citep{Tay2021} is a standard benchmark for evaluating long-context sequence models. 
It contains a suite of tasks that cover different forms of long-range reasoning, including symbolic structure processing, long-text classification, document matching, and flattened image classification.
The maximum sequence lengths range from 1k to 8k tokens depending on the task.

\paragraph{ListOps.}
ListOps tests whether a model can process hierarchical symbolic expressions over long contexts. 
Each input is a nested expression composed of operators such as maximum, minimum, median, and mean, and the model predicts the final discrete value. 
It is a ten-way classification task with a maximum input length of 2k.

\paragraph{Text.}
The Text task evaluates long-document sentiment classification. 
Given a tokenized IMDb review, the model predicts whether the review sentiment is positive or negative. 
It is a binary classification task with a maximum input length of 4k.

\paragraph{Retrieval.}
The Retrieval task evaluates whether a model can produce useful representations for matching long documents. 
Given a pair of tokenized documents from the AAN corpus, the model predicts whether a citation relation exists between them. 
It is a binary classification task with a maximum input length of 8k.

\paragraph{Image.}
The Image task evaluates whether a sequence model can recover spatial structure from a flattened image sequence. 
Each image is serialized into a one-dimensional input, and the model predicts the image class. 
It is a ten-way classification task with a maximum input length of 1k.

\subsubsection{Implementation details}

We evaluate GRIL on the LRA benchmark using the architecture shown in Figure~\ref{appendix:fig:framework}, with initialization following~\citet{orvieto2023resurrecting}. All models were trained using the AdamW optimizer with a cosine learning rate schedule and a linear warmup phase. We use early stopping based on validation performance. For each LRA task, training was limited to 24 hours using 4 NVIDIA H100 GPUs.
Specific architectural configurations and training hyperparameter settings are summarized in Table~\ref{appendix:tab:lra-model-size} and Table~\ref{appendix:tab:lra-hp}.

\begin{table*}[tbp]
\centering
\caption{Model architectural configurations across different benchmarks.}
\label{appendix:tab:lra-model-size}
\small
\begin{tabular}{lccccccc}
\toprule
\textbf{Task} & \textbf{Layers} & \textbf{Hidden} & \textbf{SSM Blocks} & \textbf{SSM State} & \textbf{Window} & \textbf{Stride} & \textbf{Params} \\
\midrule
Retrieval & 4 & 128 & 16 & 256 & 3 & 1 & 1.00M \\
ListOps       & 6 & 128 & 8  & 32  & 3 & 1 & 0.26M \\
Text   & 6 & 256 & 16 & 128 & 3 & 1 & 1.43M \\
Image  & 7 & 512 & 2  & 256 & 3 & 1 & 6.46M \\
\bottomrule
\end{tabular}
\end{table*}

\begin{table*}[tbp]
\centering
\caption{Training hyperparameters and regularization settings.}
\label{appendix:tab:lra-hp}
\small
\begin{tabular}{lcccccc}
\toprule
\textbf{Task} & \textbf{Steps} & \textbf{Warmup} & \textbf{Peak LR} & \textbf{WD} & \textbf{Dropout} & \textbf{Batch Size} \\
\midrule
Retrieval & 100k   & 10k & $4\times 10^{-3}$ & 0.01 & 0.1 & 64 \\
ListOps       & 300k   & 30k & $2\times 10^{-3}$ & 0.05 & 0.0 & 32 \\
Text    & 78.2k  & 4k  & $5\times 10^{-4}$ & 0.12 & 0.1  & 32 \\
Image & 281.4k & 28k & $4\times 10^{-3}$ & 0.05 & 0.1  & 32 \\
\bottomrule
\end{tabular}
\end{table*}

Table~\ref{appendix:tab:lra-full} reports an extended comparison on LRA~\citep{Tay2021}.
We include Mamba as a recurrent baseline with input-dependent dynamics.
The Mamba and RG-LRU baseline numbers are taken from the comparison table of~\citet{terzic_2025_pdssm} and~\citet{ssmbase}, rather than from independently reproduced or fully optimized configurations, and the remaining baselines come from existing reports under potentially different evaluation protocols, so the table should be read as a set of reference comparisons.

For GRIL, the configurations in Table~\ref{appendix:tab:lra-hp} and the reporting epochs were selected based on validation accuracy, without using test performance for selection. Each configuration was then trained with three random seeds.
Table~\ref{appendix:tab:lra-seeds} lists every run.
On ListOps, Text, Retrieval and Image, all three runs converged, and the tables report their mean.
Data for external baselines are taken from previous studies, where training and evaluation protocols may differ from those in this paper.
\begin{table*}[tbp]
    \centering
    \small
    \caption{Extended Long Range Arena accuracy (\%) comparison. GRIL entries are means over three seeds.}
    \label{appendix:tab:lra-full}
    \begin{tabular}{lcccc}
        \toprule
        \textbf{Model} & \multicolumn{4}{c}{\textbf{LRA Task [\%]}} \\
        \cmidrule(lr){2-5}
        & ListOps & Text & Retrieval & Image \\
        \midrule
        Random & 10.00 & 50.00 & 50.00 & 10.00 \\
        Transformer~\citep{Tay2021}  & 36.37 & 64.27 & 57.46 & 42.44 \\
        \midrule
        S4~\citep{gu_efficiently_2021}  & 59.60 & 86.82 & 90.90 & 88.65 \\
        S4D~\citep{Gu2022s4d}  & 60.52 & 87.34 & 91.09 & 88.19 \\
        S5~\citep{smith_simplified_2022}  & \textbf{62.15} & 89.31 & \textbf{91.40} & 88.00 \\
        LRU~\citep{orvieto2023resurrecting}  & 60.20 & \textbf{89.40} & 89.90 & \textbf{89.00} \\
        Mamba~\citep{terzic_2025_pdssm}  & 38.0 & 83.0 & 72.1 & 69.8 \\
        RG-LRU~\citep{ssmbase}  & 32.34 & 71.75 & 66.58 & 61.15 \\
        \midrule
        \textbf{GRIL} & 61.58 & 86.64 & 87.68 & 71.98 \\
        \bottomrule
    \end{tabular}
\end{table*}

\begin{table}[tbp]
    \centering
    \small
    \caption{Per-run GRIL test accuracy (\%) on Long Range Arena. Each run reports the test accuracy at its epoch of highest validation accuracy.}
    \label{appendix:tab:lra-seeds}
    \begin{tabular}{lcccc}
        \toprule
        & ListOps & Text & Retrieval & Image \\
        \midrule
        Run 1 & 62.40 & 86.86 & 87.48 & 71.72 \\
        Run 2 & 60.65 & 86.47 & 87.65 & 71.66 \\
        Run 3 & 61.70 & 86.58 & 87.91 & 72.56 \\
        \midrule
        Mean $\pm$ std & $61.58 \pm 0.88$ & $86.64 \pm 0.20$ & $87.68 \pm 0.22$ & $71.98 \pm 0.50$ \\
        \bottomrule
    \end{tabular}
\end{table}

\subsection{WikiText-103}
\label{appendix:wikitext}

We evaluate language modeling on WikiText-103~\citep{merity2016pointer}, a standard benchmark constructed from Wikipedia articles. We use the publicly available S5 language-modeling branch~\citep{smith_simplified_2022} as the implementation basis for this experiment, including its WikiText-103 dataloading, tokenization, and preprocessing pipeline. This provides a standard language-modeling setup in which we instantiate GRIL as a recurrent sequence modeling layer.

Our WikiText-103 GRIL uses hidden dimension $d_{\mathrm{model}}=768$ and MLP inner dimension $d_{\mathrm{inner}}=3072$, with initialization following~\citet{orvieto2023resurrecting}.
The base model uses $18$ layers ($127.8$M parameters), matching the baselines in parameter count; we also train a $16$-layer variant (${\approx}118$M parameters) under the same recipe, changing only the layer count.
Each GRIL layer uses an SSM size of $128$ with $32$ blocks, a causal sliding window of size $w{=}10$, and stride $1$ over consecutive token representations.
We train with sequence length $1024$, a global batch size of $16$ sequences across the 4 GPUs with no gradient accumulation, cosine learning rate decay with peak learning rate $8\times 10^{-4}$, weight decay $0.1$, $20{,}000$ warmup steps, and a budget of $290{,}720$ steps.
Dropout is $0.15$ inside the GRIL layer and $0.2$ on the embedding and residual paths, and we use seed $0$.
The $18$-layer model reaches validation perplexity $17.10$ and test perplexity $17.62$.
The $16$-layer variant reaches validation PPL $17.27$ and test PPL $17.74$; its best validation loss ($2.8488$) occurs at step ${\approx}269$k and is flat thereafter, and we report its final checkpoint (validation loss $2.8491$, test loss $2.8760$).
Training took approximately $8.7$ hours on 4 NVIDIA GH200 GPUs.
The window-size controls in Appendix~\ref{sec:window-size} use the same architecture and a shorter matched budget ($218{,}040$ steps, $15$k warmup); the $18$-layer models reported here differ from those runs in depth, schedule and warmup.

We also train a variant that changes three details of the language-model mixer while keeping the windowed cross-product write and the multiplicative readout.
At initialization the mixer output was about $250\times$ larger than the token embeddings, so we initialize the output projection of each GRIL mixer with standard deviation $0.02/\sqrt{2\,n_{\mathrm{layer}}}$, following GPT-2 and Mamba~\citep{gu_mamba_2023}, and apply an RMSNorm to the mixer output.
We further make the recurrent decay input-dependent through an RG-LRU-style gate~\citep{de_griffin_2024} computed from the newest token.
The constructions of Section~\ref{sec:1dregression}--\ref{sec:nonlinear} use a fixed decay, so this gate is a language-modeling extension and is not covered by the GD analysis.
Under the same training budget, the variant reaches test perplexity $16.77$, compared with $17.62$ for the base model; this variant is the GRIL entry in Table~\ref{tab:main-wikitext-results}.

The baselines in Table~\ref{tab:main-wikitext-results} are trained under the same GPT-2 tokenizer, the same $1024$-token non-overlapping evaluation, and a matched optimizer budget (global batch $16$, peak learning rate $8\times 10^{-4}$, weight decay $0.1$, $20{,}000$ warmup steps with cosine decay, ${\approx}37$ epochs).
We compare against TTT-Linear~\citep{sun_learning_2024} ($124.4$M), Mamba-2~\citep{dao_transformers_2024} ($129.0$M, official \texttt{mamba\_ssm} kernels), and Gated DeltaNet-2~\citep{hatamizadeh_gdn2_2026} ($134.3$M, NVLabs kernels), each run inside the same training pipeline.
Perplexities are scored on non-overlapping windows; overlapping-window evaluation would read roughly $0.6$ perplexity lower on WikiText-103.

For WikiText-103, we use the standard autoregressive next-token prediction objective. 
Each training block is constructed as an input sequence $(x_1,\ldots,x_L)$ paired with the shifted target sequence $(x_2,\ldots,x_{L+1})$. 
In the GRIL layer, the local window at position $t$ contains only left-padded context up to the current input token, i.e., $(x_{t-w+1},\ldots,x_{t-1},x_t)$ for a window of size $w$. 
The corresponding prediction is evaluated against $x_{t+1}$, which lies outside the input window. 
Therefore, the windowed computation preserves the causal autoregressive structure.

For additional context, Table~\ref{appendix:tab:wikitext-full} collects previously reported WikiText-103 perplexities at a similar parameter scale.
These numbers are taken from the corresponding public repositories\footnote{\label{fn:s5}\url{https://github.com/lindermanlab/S5/tree/development}} \footnote{\url{https://github.com/state-spaces/mamba/issues/8}} and serve only as reference points; unlike the matched comparison in Table~\ref{tab:main-wikitext-results}, they come from different codebases and may use different training and evaluation protocols.

\begin{table}[tbp]
\centering
\caption{Previously reported perplexities on the \textsc{WikiText-103} test set at a similar parameter scale, included as reference points.
These come from different codebases and may use different protocols; the matched-budget comparison is Table~\ref{tab:main-wikitext-results}.}
\label{appendix:tab:wikitext-full}

\begin{tabular}{l@{\hspace{-4pt}}c}
\toprule
\textbf{Model} & \textbf{Perplexity} \\
\midrule
Transformer (125M)~\citep{fu2023hungryhungryhipposlanguage} & 18.6 \\
Hybrid H3 (125M) & 18.5 \\
Performer (125M) & 26.8 \\
Reformer (125M) & 25.6 \\
\midrule
AFT-conv (125M) & 28.2 \\
Linear SA (125M)~\citep{fu2023hungryhungryhipposlanguage} & 25.6 \\
\midrule
Hyena-Convolution (125M) & 18.6 \\
Hyena-slim-Convolution (125M) & 18.5 \\
Hyena-S5 (125M) & 18.3 \\
\midrule
\textbf{Mamba (130M)}~\citep{gu2023mamba-repo} & 16.3 \\
GRIL (118M, $w=10$) & 17.74 \\
\bottomrule
\end{tabular}
\end{table}

\input{window_size_lm}

\subsection{Long-context associative recall} \label{sec:ar-associative-recall}

We evaluate GRIL on the AR setting introduced by
Hyena~\citep{hyena}: sequence length $131{,}072$ and vocabulary size $30$.
In this task, each sequence concatenates many key--value pairs drawn from a random
dictionary; the model must return the value associated with a queried key that
appears earlier in the context.
Following the Hyena-S5 reproduction setup\footref{fn:s5},
we train two-layer language-model models with width $d{=}64$ and MLP width $256$
for $50{,}000$ steps (cosine schedule, peak learning rate $10^{-3}$, weight decay
$0.1$, effective batch size $16$).
For GRIL we use state size $32$ and a causal local window of size $w{=}10$.
We report last-token test accuracy.
Each seed took at most $6$ hours on 4 NVIDIA A100-SXM4-40GB GPUs.

The comparison is reported in the main text in Table~\ref{tab:ar-131k-30}; here we give the full setup and the source of each number.
Hyena with convolutions reaches $97.2\%$ accuracy~\citep{hyena}, and Hyena-S5 is the $10$-seed mean of $97.2\%$ from the S5 development branch. Mamba achieves $99.5\%$ accuracy, and GRIL attains the best result with a $10$-seed mean accuracy of $99.9\%$, with individual runs ranging from $99.8\%$ to $100.0\%$. The Transformer, FlashTransformer, GSS, H3, AFT, and RWKV baselines are as reported by \citet{hyena}: the two attention variants do not fit the $131$K-token context ($\ding{55}$), and GSS, H3, AFT, and RWKV reach $0.1\%$, $0.6\%$, $0.8\%$, and $2.3\%$ respectively. Hyena-S5, Mamba, and GRIL use the same outer LM architecture and training budget ($d=64$, $n=2$, MLP width $256$). GRIL additionally uses a short causal window of $w=10$ and an SSM state size of $32$.

\section{Relation to other recurrent formulations} \label{appendix:longhorn-perspective}

We compare GRIL with linear attention, Longhorn, and delta-rule recurrent models, and then relate depth to looped computation.

\subsection{Relationship with linear self-attention}\label{appendix:lsa}

GRIL instead uses the weighted local outer-product update in Eq.~\ref{eq:window-cross-product} in place of the single-token outer product of linear self-attention (Table~\ref{tab:recurrent-comparison}).
The update is followed by a readout analogous to linear self-attention.
The recurrent write is lifted from a single-token product to a window-level product that can select cross-token supervised factors.
This view explains why a sliding-window variant of linear attention can implement gradient-descent-style ICL with learned recurrent degrees of freedom that scale as $\mathcal{O}(f^2)$ for an $f\times f$ linear-regression task.
By contrast, the gated-recurrent emulation of linear attention analyzed by \citet{zucchet_2023_gated} can require $\mathcal{O}(f^4)$ parameters in the same setting.
The saving comes from exposing the supervised cross-products directly in the local window, rather than representing them through a generic recurrent implementation of linear attention.

Section~\ref{sec:ndregression} shows that the $w=1$ limit collapses this recurrence to the single-token form, which removes the temporal structure a one-layer supervised-GD reading requires: a single token cannot simultaneously expose $\vx_t$, $\vy_t$, and $\vx_{t+1}$.
A $w=1$ model therefore cannot realize the same one-layer construction directly and must assemble the required cross-token interactions through additional recurrent computation.
Table~\ref{tab:synthetic-window} reports the corresponding sweep on multivariate linear regression, where $w{=}3$ recovers the analytic one-step GD loss and the other windows tested fail to recover it.
Windows wider than three still contain the required triple, so their loss reflects what training finds rather than what the architecture can represent (Appendix~\ref{sec:window-size-synthetic}).

\subsection{Windowed projections and short convolutions} \label{appendix:shortconv}

The windowed query, key, and value embeddings of Section~\ref{sec:ndregression}, whose columns are the token-wise projections over the local window, place GRIL in the same family as recurrent models that read their query, key, and value vectors from a window rather than from a single token.
In linear-attention and state-space models such as Mamba~\citep{gu_mamba_2023}, DeltaNet~\citep{yang_parallelizing_2024}, and Gated DeltaNet~\citep{yang_gated_2025}, forming these windows is a short depthwise causal convolution over the projections, and BASED~\citep{arora_simple_2024} places a short convolution beside its linear-attention state.
A one-hot convolution kernel is the special case that selects a single window position, which is a fixed shift.

\paragraph{The construction as a one-hot convolution.}
The one-step GD construction uses one-hot selectors that read different offsets on the write and read paths.
On the read path, $\vc=\ve_3$ gives the most recent column $\mQ_t\vc=\vq_t$.
When the window-mixing matrix is rank one, $\mC=\ve_2\ve_1^\top$, the write factors as
\[
    \mV_t\mC\mK_t^\top=(\mV_t\ve_2)(\mK_t\ve_1)^\top=\vv_{t-1}\vk_{t-2}^\top,
\]
pairing the value at offset $t-1$ with the key at offset $t-2$.
In the construction these projections are the identity, so the columns are the raw window tokens $\vk_{t-2}=\vx_t$, $\vv_{t-1}=\vy_t$, and $\vq_t=\vx_\tpone$, and the write and readout reduce to $\vy_t\vx_t^\top$ and $\mZ_t\vx_\tpone$ (Section~\ref{sec:ndregression}).
This is short-convolution linear attention in which the value and key are read by one-hot convolutions at different shifts, applied over the stride-two interleaved sequence.
The second multi-step head factors the same way, with $\mC=\ve_1\ve_1^\top$ giving $\vv_{t-2}\vk_{t-2}^\top=\vx_t\vx_t^\top$ (Appendix~\ref{appendix:multi}).

\paragraph{What the general block adds.}
GRIL keeps $\mC$ as a full $w\times w$ matrix, so the write
\[
    \mV_t\mC\mK_t^\top=\sum_{i,j}C_{ij}\,\vv_i\vk_j^\top
\]
is a weighted sum of outer products across window positions.
A single product $\vv_{t-1}\vk_{t-2}^\top$ of one-hot-convolved factors cannot represent this sum once $\mC$ has rank above one, so the block is strictly more expressive than its short-convolution special case.
The one-hot convolution is the particular point within this family that the GD construction singles out.
The read path stays a vector $\vc\in\R^w$, so $\vq_t=\mQ_t\vc$ is a windowed query for every $\vc$, and a non-one-hot $\vc$ reads a learned mixture of window positions instead of a single shift.

\paragraph{What separates the construction from a generic short convolution.}
Two properties distinguish the GD-implementing configuration from a standard short-convolution model, even though both read $\vq_t$, $\vk_t$, and $\vv_t$ from a window.
First, the stride-two interleaving aligns one training example with each recurrent step.
A position-independent stride-one kernel that selects $(\vx_t,\vy_t)$ on the aligned window also fires on the misaligned window containing $(\vy_{t-1},\vx_t,\vy_t)$, while the stride-two window advances one example per step and keeps every write aligned.
Second, trained short-convolution and delta-rule baselines do not recover the construction on the synthetic task, while the trained GRIL does (Appendix~\ref{sec:deltanet-baselines}).
The windowed projections therefore place GRIL inside the short-convolution family, and the one-step GD solution is a specific one-hot, stride-two, cross-offset configuration within it that training is biased toward.

\subsection{Longhorn's online-optimization view} \label{appendix:longhorn-view}

Longhorn frames a sequence-mixing SSM as an amortized online learner: the recurrent state is updated as the solution of an online learning objective, giving the transition rule an optimization-derived form~\citep{liu_longhorn_2024}. In their online regression example, the state $\mS_t$ is chosen to stay close to the previous state while predicting the current token $\vx_t$ from a key $\vk_t$, with tradeoff coefficient $\beta_t>0$:
\begin{align*}
    \mS_t
    = \arg\min_{\mS}
    \left\|\mS-\mS_{t-1}\right\|_F^2
    + \beta_t \left\|\mS \vk_t-\vx_t\right\|_2^2 .
\end{align*}
The corresponding implicit update can be written, up to orientation conventions, as
\begin{align*}
    \mS_t
    = \mS_{t-1}
    + \Delta_t(\vx_t-\mS_{t-1}\vk_t)\vk_t^T, \\
    \Delta_t = \frac{\beta_t}{1+\beta_t\|\vk_t\|_2^2}.
\end{align*}
Thus Longhorn is an online associative-memory learner: the state is edited so that key $\vk_t$ recalls value $\vx_t$, and the closed-form implicit step supplies a stable forgetting/update factor.

Our GRIL construction fits this optimization perspective, but with a different objective and a different local information pattern. In Section~\ref{appendix:lsa}, we show that the $w=1$ limit reduces to the same broad single-token outer-product memory family as linear attention. The full GRIL construction, however, uses $w=3$, so the value and key embeddings of the window are
\[
    \mV_t = \mK_t = \mQ_t = [\vx_t,\; \vy_t,\; \vx_{t+1}].
\]
This lets one recurrent step access the supervised training pair $(\vx_t,\vy_t)$ and the next query $\vx_{t+1}$ simultaneously. The update
\begin{align*}
    \mZ_t = \mZ_{t-1} + \mV_t\mC\mK_t^T
\end{align*}
therefore accumulates one contribution to the supervised least-squares gradient per recurrent step, because choosing $\mC$ appropriately yields the term $\vy_t\vx_t^T$ when $\mW_0=\vzero$. No update is applied at this point. The readout $\mZ_t\mQ_t\vc$ takes the gradient step, using the statistic accumulated over the whole context so far, and applies it to $\vx_{t+1}$.

The distinction from Longhorn is therefore both the objective being amortized and the schedule on which it is applied.
Longhorn chooses an online associative-recall objective and uses its implicit closed-form update to obtain a stable state transition, editing the state once per example.
GRIL instead chooses a supervised objective whose update target is the gradient over the in-context examples, and defers the update to the readout, so the step it takes is a mini-batch step over the whole context rather than a per-example one.
The sliding window is what exposes the supervised quantities needed for that objective in one recurrent step.

\subsection{Online versus mini-batch updates}
\label{appendix:online-vs-minibatch}

A second, closely related distinction is \emph{when} a supervised gradient is applied, which Section~\ref{sec:ndregression} motivates for the construction.
To make this concrete without confounding neural parameterization, we evaluate non-neural baselines on the same synthetic multivariate linear-regression episodes used elsewhere (input dimension $10$, $10$ context examples, $1000$ episodes).
Both online GD and mini-batch GD are tuned over the same learning-rate grid.
Table~\ref{tab:online-vs-minibatch} reports query MSE.

\begin{table}[!t]
\centering
\small
\caption{Non-neural baselines on synthetic vector linear regression
(dim.\ $10$, $10$ context examples, $1000$ episodes).
Online GD and mini-batch GD use the same number of updates and
learning-rate grid.}
\label{tab:online-vs-minibatch}

\setlength{\tabcolsep}{2pt}
\begin{tabular}{
@{}
p{0.24\columnwidth}
p{0.30\columnwidth}
p{0.16\columnwidth}
p{0.22\columnwidth}
@{}
}
\toprule
\textbf{Method} &
\textbf{Update rule} &
\textbf{Updates} &
\textbf{Query MSE} \\
\midrule

Zero predictor &
No adaptation &
0 &
$0.411 \pm 0.220$ \\

Least squares &
Closed-form full-context &
Oracle &
$0.000 \pm 0.001$ \\

Mini-batch GD &
Full-context gradient &
10 &
$0.105 \pm 0.092$ \\

Online GD &
Per-example gradient, one pass &
10 &
$0.160 \pm 0.116$ \\

\bottomrule
\end{tabular}
\end{table}

Even with a matched update count, online GD remains a noisier baseline than mini-batch GD in this supervised ICL setting ($0.160$ vs.\ $0.105$).
This is not a claim that mini-batch GD universally dominates online fast-weight methods; together with the Longhorn comparison above, it clarifies why GRIL targets a supervised mini-batch gradient statistic rather than per-example associative or online updates.

\subsection{Matched DeltaNet and Gated DeltaNet baselines}

\label{sec:deltanet-baselines}

DeltaNet~\citep{yang_parallelizing_2024} and Gated DeltaNet~\citep{yang_gated_2025} are neural instances of the same online family, because both edit the recurrent state with a per-example delta-rule correction instead of accumulating a full-context gradient statistic.
Table~\ref{tab:online-vs-minibatch} separates the two update schedules without a neural parameterization, so it does not determine whether a trained delta-rule model with comparable local mixing recovers the GD construction on its own.
We therefore evaluate both models on the same synthetic multivariate linear-regression task, sweeping the short-convolution width $\mathrm{conv}\in\{2,3,4\}$, learning rate $\mathrm{lr}\in\{10^{-4},3{\times}10^{-4},5{\times}10^{-4},10^{-3}\}$, and depth $\in\{1,4\}$.
A width-3 short convolution puts the same three tokens in the receptive field that GRIL's $w{=}3$ window exposes, so the comparison isolates the effect of \emph{how} those tokens enter the recurrent write and read paths.
Two differences remain under matched access.
The convolution collapses its receptive field into one vector by a weighted sum before anything bilinear happens, so recovering the term $\vy_t\vx_t^T$ exactly requires the key and value kernels to be one-hot selectors of two specific positions, whereas GRIL keeps the tokens as separate columns of $\mV_t$ and $\mK_t$ and performs the selection in $\mC$, after the window rather than inside it.
The convolution kernels are also position-independent while the delta rule writes at every step, so a kernel that selects $(\vx_t,\vy_t)$ on the aligned window also fires on the misaligned one containing $(\vy_{t-1},\vx_t,\vy_t)$, whereas GRIL's stride of two advances one training example per recurrent step and keeps every write aligned.
Table~\ref{tab:deltanet-baselines} reports the best validation MSE per configuration under this matched grid.

\begin{table}[t]
\centering
\small
\caption{Matched DeltaNet and Gated DeltaNet baselines on synthetic multivariate linear regression, against exact one-step GD and GRIL.
Best validation MSE over $\mathrm{lr}\in\{10^{-4},3{\times}10^{-4},5{\times}10^{-4},10^{-3}\}$.
Lower loss means a better predictor, but only a value matching $0.207$ reproduces the analytic one-step GD solution.}
\label{tab:deltanet-baselines}
\begin{tabular}{llcc}
\toprule
\textbf{Model} & \textbf{Depth} & \textbf{Local size} & \textbf{Loss} \\
\midrule
Exact one-step GD & -- & -- & $0.207$ \\
\textbf{GRIL} & 1 & $w{=}3$ & $\mathbf{0.206}$ \\
\midrule
DeltaNet & 1 & $\mathrm{conv}{=}2$ & $0.158$ \\
DeltaNet & 1 & $\mathrm{conv}{=}3$ & $0.155$ \\
DeltaNet & 1 & $\mathrm{conv}{=}4$ & $0.154$ \\
Gated DeltaNet & 1 & $\mathrm{conv}{=}2$ & $0.152$ \\
Gated DeltaNet & 1 & $\mathrm{conv}{=}3$ & $0.154$ \\
Gated DeltaNet & 1 & $\mathrm{conv}{=}4$ & $0.155$ \\
\midrule
DeltaNet & 4 & $\mathrm{conv}{=}3$ & $0.031$ \\
DeltaNet & 4 & $\mathrm{conv}{=}4$ & $0.030$ \\
Gated DeltaNet & 4 & $\mathrm{conv}{=}3$ & $0.040$ \\
\bottomrule
\end{tabular}
\end{table}

Both baselines reach $0.152$--$0.158$ at depth 1 and $0.030$--$0.040$ at depth 4, well below the exact one-step GD loss of $0.207$, but none of them lands on the GD value at any convolution width or depth.
GRIL at $w{=}3$ is the only model in Table~\ref{tab:deltanet-baselines} that reproduces the analytic one-step GD loss ($0.206$ vs.\ $0.207$), and Figure~\ref{fig:lossnmatrices} shows that it also recovers the parameters predicted by the construction.
The delta-rule baselines are therefore stronger predictors on this task, while GRIL is the architecture whose trained solution coincides with the analytic learning algorithm.
Because the benchmark here is one-step GD itself, reaching a loss below the GD value is evidence that a model implements a different, more aggressive update rule, not that it recovers the target construction; it is the coincidence with the GD loss and parameters, not the raw loss, that the synthetic task is designed to measure.
A width-3 short convolution does not by itself recover the construction, which supports the interpretation in Section~\ref{appendix:lsa} that the GD solution is identified by the separation between the write path ($\mC$ selects $\vy_t\vx_t^\top$) and the read path ($\vc$ enters only at readout) rather than by access to a three-token window.

\clearpage

\end{document}

%% file: math_commands.tex
\usepackage{amsmath,amsfonts,bm}

\def\eqref#1{Eq.~\ref{#1}}

\def\1{\bm{1}}

\def\vzero{{\bm{0}}}

\def\va{{\bm{a}}}
\def\vb{{\bm{b}}}
\def\vc{{\bm{c}}}

\def\ve{{\bm{e}}}

\def\vg{{\bm{g}}}

\def\vk{{\bm{k}}}

\def\vo{{\bm{o}}}
\def\vp{{\bm{p}}}
\def\vq{{\bm{q}}}

\def\vs{{\bm{s}}}

\def\vv{{\bm{v}}}
\def\vw{{\bm{w}}}
\def\vx{{\bm{x}}}
\def\vy{{\bm{y}}}
\def\vz{{\bm{z}}}

\def\mA{{\bm{A}}}
\def\mB{{\bm{B}}}
\def\mC{{\bm{C}}}

\def\mE{{\bm{E}}}

\def\mI{{\bm{I}}}

\def\mK{{\bm{K}}}

\def\mQ{{\bm{Q}}}

\def\mS{{\bm{S}}}

\def\mU{{\bm{U}}}
\def\mV{{\bm{V}}}
\def\mW{{\bm{W}}}

\def\mZ{{\bm{Z}}}

\def\mPsi{{\bm{\Psi}}}

\def\mTheta{{\bm{\Theta}}}

\DeclareMathAlphabet{\mathsfit}{\encodingdefault}{\sfdefault}{m}{sl}
\SetMathAlphabet{\mathsfit}{bold}{\encodingdefault}{\sfdefault}{bx}{n}

\newcommand{\R}{\mathbb{R}}

\renewcommand{\L}{\ensuremath{\mathcal{L}}}
\newcommand{\D}{\ensuremath{\mathcal{D}}}
\newcommand{\tpone}{\ensuremath{{t+1}}}
\newcommand{\hy}{\ensuremath{\hat{y}}}
\newcommand{\hvy}{\ensuremath{\hat{\vy}}}

%% file: window_size.tex
\subsection{Window size}
\label{sec:window-size-synthetic}

The constructions in Sections~\ref{sec:1dregression}--\ref{sec:ndregression} identify $w{=}3$ as the \emph{minimal} local window that places the current input, current target, and next query in one recurrent step.
Table~\ref{tab:synthetic-window} tests this prediction on the same multivariate linear-regression episodes used elsewhere (exact one-step GD loss $0.207$).
Under a matched one-layer GRIL recipe, $w{=}3$ recovers the analytic GD loss ($0.206$), while $w\in\{1,2,4,5\}$ fail to recover it and stay well above the GD solution ($0.668$--$0.836$).
The two sides of this sweep fail for different reasons.
Windows below three cannot expose $\vx_t$, $\vy_t$ and $\vx_{t+1}$ in one recurrent step, so the construction rules them out, whereas windows above three contain the same triple as a sub-window, and a larger $\mC$ could in principle select the two relevant positions and ignore the rest.
The loss at $w\in\{4,5\}$ is therefore evidence about optimization rather than about representational capacity, and it indicates that the minimal window supplies an inductive bias toward the intended factorization rather than that wider windows cannot express it.

\begin{table}[t]
\centering
\small
\caption{Synthetic multivariate linear regression window sweep (one-layer GRIL).
Best validation MSE; exact one-step GD obtains $0.207$.}
\label{tab:synthetic-window}
\begin{tabular}{lcc}
\toprule
\textbf{Model} & \textbf{Window $w$} & \textbf{Loss} \\
\midrule
Exact one-step GD & -- & $0.207$ \\
GRIL & 1 & $0.835$ \\
GRIL & 2 & $0.668$ \\
\textbf{GRIL} & $\mathbf{3}$ & $\mathbf{0.206}$ \\
GRIL & 4 & $0.836$ \\
GRIL & 5 & $0.836$ \\
\bottomrule
\end{tabular}
\end{table}

That argument is specific to the supervised ICL tokenization and does not imply that $w{=}3$ is optimal for every sequence-modeling objective, which Appendix~\ref{sec:window-size} examines under a language-modeling objective.

%% file: multi_layer_depth.tex
\subsection{Multi-step GRIL: setup and depth study}
\label{sec:multi-layer-depth}

For the multi-step regression setting, we use a multi-layer GRIL architecture without non-linearities between layers.
For the explicit construction, a $k$-layer model implements $k$ gradient-descent steps.
All hyper-parameters are kept the same as in the single-layer setting.
We compare two variants: (i) a fully trained model (denoted GRIL), and (ii) an explicit linear GD construction.

The explicit construction fixes all recurrent parameters to the values derived in Appendix~\ref{appendix:multi}.
Following~\citet{von_oswald_transformers_2023}, we optimize only the step size $\eta_l$ of each layer on the regression objective, while keeping all other constructed parameters fixed.
The one-step constructive validation uses its single step size directly.

Figure~\ref{fig:result-regression-multi-layer}A reports the two-layer case.
We extend the experiment to three and four layers and compare GRIL with matched multi-step GD and GD$^{++}$~\citep{von_oswald_transformers_2023}.
GD$^{++}$ additionally applies the input transformation
\[
\vx_i \leftarrow \bigl(I-\gamma XX^\top\bigr)\vx_i\,,
\]
where $X$ stacks the in-context inputs and $\gamma$ controls the correction.

\begin{table}[tbp]
\centering
\small
\caption{Multi-layer GRIL on multivariate linear regression versus matched multi-step GD and GD$^{++}$. Losses are validation MSE. Absolute gaps are reported relative to each reference.}
\label{tab:multi-layer-depth}

\begin{tabular}{cccccc}
\toprule
Layers & GRIL & GD & $\Delta_{\mathrm{GD}}$ & GD$^{++}$ & $\Delta_{\mathrm{GD}^{++}}$  \\
\midrule
2 & 0.141 & 0.156 & 0.015 & 0.108 & 0.033 \\
3 & 0.111 & 0.132 & 0.021 & 0.088 & 0.023 \\
4 & 0.092 & 0.118 & 0.025 & 0.076 & 0.016 \\
\bottomrule
\end{tabular}
\end{table}

Across the evaluated depths, GRIL validation loss decreases from $0.141$ at two layers to $0.111$ at three layers and $0.092$ at four layers.
GRIL remains below matched plain GD and above GD$^{++}$, while the gap to GD$^{++}$ decreases from $0.033$ to $0.016$.
We evaluate depths only up to four layers.

%% file: window_size_lm.tex
\subsubsection{Window size}
\label{sec:window-size}

For next-token prediction there is no analogous closed-form argument, so we ablate the window directly under a matched language-modeling budget.
We train approximately equal-capacity GRIL language models ($\approx$118M parameters; $d{=}768$, $16$ layers, SSM size $128$) on WikiText-103, varying only the local window size.
All runs in Table~\ref{tab:wiki-window} share the same optimizer and budget: learning rate $8{\times}10^{-4}$ with $15$k warmup and cosine decay, dropout $0.15$/$0.2$, and $218{,}040$ training steps, and each window is trained with three seeds.
Changing the window perturbs the parameter count by at most $0.04\%$.

\begin{table}[t]
\centering
\small
\caption{WikiText-103 window ablation under a matched $\approx$118M GRIL recipe and a fixed $218$k-step budget.
Each entry is the mean $\pm$ one standard deviation over three seeds; perplexity is $\exp(\text{loss})$.}
\label{tab:wiki-window}
\begin{tabular}{lccccc}
\toprule
\textbf{Window $w$} & \textbf{\#Params} & \textbf{Val.\ loss} & \textbf{Val.\ PPL} & \textbf{Test loss} & \textbf{Test PPL} \\
\midrule
3   & 117.90M & $2.84 \pm 0.018$ & $17.12 \pm 0.30$ & $2.87 \pm 0.016$ & $17.64 \pm 0.29$ \\
10  & 117.90M & $2.87 \pm 0.016$ & $17.64 \pm 0.28$ & $2.89 \pm 0.020$ & $18.16 \pm 0.36$ \\
50  & 117.94M & $2.96 \pm 0.015$ & $19.30 \pm 0.25$ & $2.99 \pm 0.016$ & $19.88 \pm 0.28$ \\
\bottomrule
\end{tabular}
\end{table}

At this budget, $w{=}3$ attains the lowest loss on both splits (validation loss $2.84 \pm 0.018$, test perplexity $17.64 \pm 0.29$), ahead of $w{=}10$ by $0.03$ in validation loss and $0.52$ in test perplexity.
A wide window clearly hurts: $w{=}50$ raises the validation loss to $2.96 \pm 0.015$ and the test perplexity to $19.88 \pm 0.28$, more than five standard deviations above $w{=}10$.
A short window also helps long-context language modeling in hybrid sliding-window architectures, where widening the window degrades long-range performance because the model then leans less on its recurrent memory~\citep{cabannes_short_2025}.
The parameter-matched comparison in Table~\ref{tab:main-wikitext-results} uses an $18$-layer GRIL with $w{=}10$ trained for the longer $290$k-step schedule (Appendix~\ref{appendix:wikitext}); we did not train the $w{=}3$ setting at that depth and budget.